\pdfoutput=1
\documentclass[11pt]{article}

\usepackage[]{EMNLP2023}

\usepackage{times}
\usepackage{latexsym}

\usepackage{inconsolata}

\usepackage{blindtext}
\usepackage{amsmath}
\usepackage{booktabs}
\usepackage{lscape} 
\usepackage{graphicx}
\usepackage{subcaption}
\usepackage{xspace}
\usepackage{arydshln}
\usepackage{bbm}
\usepackage[normalem]{ulem}
\usepackage{multirow}
\usepackage{tabularx}
\usepackage{todonotes}
\usepackage{amssymb}
\usepackage[acronym]{glossaries}
\usepackage{makecell}
\usepackage{textcomp}
\usepackage{rotating,amsmath,booktabs}

\newcommand{\hnc}{\textsc{HNC}\xspace}
\newcommand{\hncF}{\textsc{HNC}$_\text{full}$\xspace}
\newcommand{\hncS}{\textsc{HNC}$_\text{subset}$\xspace}
\newcommand{\foil}{\textsc{FOIL}\xspace}

\newcommand{\cwwv}{\textsc{CWWV}$_{Img}$\xspace}

\newcommand{\volta}{\textsc{Volta}\xspace}

\newcommand{\uniter}{\textsc{UNITER}\xspace}

\newcommand{\vilbert}{\textsc{VilBERT}\xspace}

\newcommand{\lxmert}{\textsc{LXMERT}\xspace}

\newcommand{\visualbert}{\textsc{VisualBERT}\xspace}

\newcommand{\vlbert}{\textsc{VL-BERT}\xspace}

\usepackage[T1]{fontenc}
\usepackage[utf8]{inputenc}

\usepackage{microtype}

\title{HNC: Leveraging Hard Negative Captions towards Models with Fine-Grained Visual-Linguistic Comprehension Capabilities}

\author{Esra Dönmez$^\ast$, Pascal Tilli$^\ast$, Hsiu-Yu Yang$^\ast$, Thang Vu, Carina Silberer \\ 
        Institute for Natural Language Processing, University of Stuttgart \\
        \small{\texttt{\{esra.doenmez, pascal.tilli, hsiu-yu.yang, thang.vu, carina.silberer\} @ims.uni-stuttgart.de}}}
        
\begin{document}

\newacronym{vl}{VL}{Vision and Language}
\newacronym{vlm}{VLM}{Vision and Language Model}
\newacronym{vlms}{VLMs}{Vision and Language Models}
\newacronym{itm}{ITM}{Image-Text-Matching}
\newacronym{mlm}{MLM}{Masked Language Modeling}
\newacronym{lm}{LM}{Language Model}
\newacronym{mrm}{MRM}{Masked Region Modeling}
\newacronym{hnc}{HNC}{Hard Negative Captions}
\newacronym{volta}{VOLTA}{Visiolinguistic Transformer Architectures}
\newacronym{valse}{VALSE}{\textbf{V}ision \textbf{A}nd \textbf{L}anguage \textbf{S}tructured \textbf{E}valuation}
\newacronym{cpt}{CPT}{\textbf{C}ommonsense \textbf{P}robing \textbf{T}ask}
\newacronym{gqa}{GQA}{Visual Reasoning in the Real World}
\newacronym{pos}{PoS}{Part-of-Speech}
\newacronym{covr}{COVR}{\textbf{CO}mpositional \textbf{V}isual \textbf{R}easoning}
\newacronym{blip}{BLIP}{Bootstrapping Language-Image Pre-training for unified vision-language understanding and generation}

\maketitle
\def\thefootnote{*}\footnotetext{These authors contributed equally to this work.}\def\thefootnote{\arabic{footnote}}
\begin{abstract}
Image--Text-Matching (ITM) is one of the defacto methods of learning generalized representations from a large corpus in Vision and Language (VL). 
However, due to the weak association between the web-collected image--text pairs, models fail to show a fine-grained understanding of the combined semantics of these modalities.
To address this issue we propose Hard Negative Captions (HNC): an automatically created dataset containing \emph{foiled} \textbf{hard negative} captions for ITM training towards achieving fine-grained cross-modal comprehension in VL. 
Additionally, we provide a challenging manually-created test set for benchmarking models on a fine-grained cross-modal mismatch task with varying levels of compositional complexity. 
Our results show the effectiveness of training on HNC by improving the models' zero-shot capabilities in detecting mismatches on diagnostic tasks and performing robustly under noisy visual input scenarios. 
Also, we demonstrate that HNC models yield a comparable or better initialization for fine-tuning. 
Our code and data are publicly available.\footnote{https://github.com/DigitalPhonetics/hard-negative-captions under MIT License.}
\end{abstract}

\section{Introduction}
\label{introduction}
Pre-trained \gls{vlms} \cite{Su2020VL-BERT, lu2019vilbert, chen2020uniter, tan2019lxmert}, when fine-tuned on downstream tasks, show promising performance thanks to their learned generalized information (or even knowledge) \cite{zhang-etal-2019-ernie, Gan_undated-zb, hendricks-nematzadeh-2021-probing}. These models are typically trained on a combination of several datasets under self-supervised training objectives, such as \gls{itm}, \gls{mlm}, and \gls{mrm}.
\gls{itm} defines the objective of predicting whether the textual and visual modalities entail one another. 
To learn this entailment, for already weakly-associated image--caption pairs, the negative captions are typically sampled from %the
mini-batch training data which results in negative captions that do not align with the image, i.e.,~the mismatch between the modalities can be detected easily since the images and captions are semantically unrelated.
Consequently, the compositional understanding capabilities of \gls{vlms} are rather limited, e.g.,~they tend to show weaknesses in correctly grounding linguistic concepts in their visual counterparts \cite{bitton-etal-2021-automatic, Keysers2019-zk, bogin-etal-2021-covr}.
These \gls{vlms}, when tested against foiled inputs, fail against \textit{fine-grained} mismatches in multimodal data (vision and language) \cite{Shekhar2017-nr, hendricks-nematzadeh-2021-probing}. 

To address the aforementioned limitations we focus on improving \gls{vlms} by automatically creating a dataset that enables learning from hard negative captions, i.e., negative captions that are minimally contradictory to their corresponding images.
We state the hypothesis that such hard negative captions increase the general comprehension capabilities of pre-trained \gls{vlms}. 
We summarize our contributions as follows:
\begin{figure*}
    \centering
    \resizebox{1.0\textwidth}{!}{
        \includegraphics{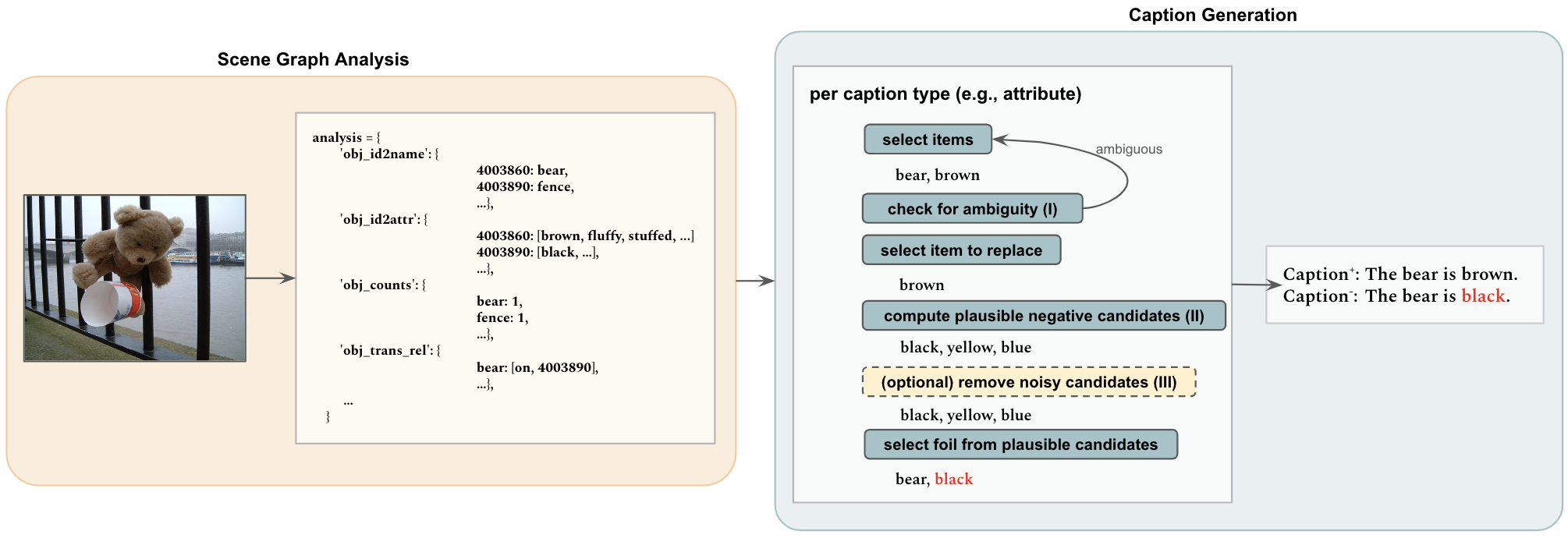}
    }
    \caption{\protect An illustration of our caption generation procedure. For each scene graph (that belongs to exactly one image) we run through this pipeline to generate hard negative captions. Details on the modules marked with Roman letters (I, II, and II) can be found in Sec. \ref{sec:datasets}.}
    \label{fig:workflow}
\end{figure*}

\begin{enumerate}
    \item We introduce \textbf{H}ard \textbf{N}egative \textbf{C}aptions (HNC) for \gls{itm} training with systematically created hard negatives: $12$ linguistically-motivated types of captions\footnote{Our code allows everyone to easily add new caption types.} that locally describe an image with their hard negative counterparts that are minimally contradictory to the given image.
    
    \item To the best of our knowledge, we are the first to \textbf{leverage scene graph information} \citep{DBLP:journals/ijcv/KrishnaZGJHKCKL17} for automatically creating hard negative captions (fine-grained misaligned image--text pairs) for \gls{itm} training. This enables us to control (1) the semantics of the hard negatives with multiple mismatch types, and (2) the level of compositional complexity in fine-grained mismatches. Our method is resource-lean in constructing the hard negatives, and flexible in that it can be extended to other phenomena which is necessary for this fast-developing \gls{vl} research field. 
    
    \item We propose a challenging human-annotated test set to benchmark \gls{vl} models' capabilities on several skills and levels of compositional understanding.
    
    \item We perform an extensive study across various tasks and show models' improvement in fine-grained cross-modal comprehension in zero-shot settings. Additionally, we show that models further trained on \gls{hnc} can serve as a better initialization point for downstream task fine-tuning.
\end{enumerate}

\section{Related Work}
\label{sec:related_work}
\paragraph{Probing \gls{vlms} for fine-grained visual grounding} 
Several works revealed shortfalls in visual grounding capabilities of \gls{vlms} at various levels by creating foiled visual descriptions in which they alter the nouns \cite{shekhar2017foil}, words belonging to other \gls{pos} tags such as adjectives or adverbs \cite{shekhar-etal-2017-vision}, S(ubject)--V(erb)--O(object) triples \cite{hendricks-nematzadeh-2021-probing}, person entities \cite{park-etal-2022-exposing}. These studies collectively suggest that \gls{vlms} struggle with fine-grained image--caption matching. Moreover, several works studied the compositional understanding of \gls{vlms} in visual grounding. 
\citet{DBLP:conf/cvpr/Winoground_Thrush22} propose Winoground to evaluate visual grounding robustness using captions with the same set of words but different syntactic structures. Their findings suggest that \gls{vlms} exhibit bag-of-words behavior \cite{diwan-etal-2022-winoground}.    
\citet{bogin-etal-2021-covr} introduce \gls{covr} to examine models' compositional generalization on unseen logical operations, e.g., quantifiers or aggregations, and conclude that reasoning over complex structures remains challenging.
While above works aim to create probing datasets to identify \gls{vlms}' potential shortfalls in visual grounding, our research goal goes beyond that: we propose a creation method for large-scale \gls{itm} datasets, useful for further pretraining (or fine-tuning) models towards fine-grained cross-modal comprehension abilities. 

\paragraph{Addressing shortfalls in fine-grained visual grounding capabilities of \gls{vlms}}
Given that \gls{vlms} are usually pre-trained with web-crawled weakly-aligned image--caption pairs, e.g., Conceptual Captions \cite{sharma-etal-2018-conceptual}, their ability to address cross-modal misalignments is questionable. 
The aforementioned empirical probes support this claim and suggest that \gls{vlms} tend to suffer from overprediction in that they consider a somewhat related image--caption pair to be associated. 
Previous works address this issue as a part of the training strategy \cite{liu-ye-2019-strong, Zhou2020MoreGI, Chen2020-oe, chen-etal-2022-contrastive}, the model architecture \cite{Messina2020TransformerReasoning, DBLP:journals/mta/ZhangHQLL22_CMRN}, or by augmenting training data \cite{shekhar2017foil, DBLP:conf/bmvc/VSE++_FaghriFKF18, DBLP:conf/eccv/GuptaVC0KH20}. 
We contribute to the last line of research and propose to augment hard negative captions for \gls{itm} training by leveraging scene graphs towards achieving a fine-grained \gls{vl} comprehension.

\section{\gls{hnc}: Hard Negative Captions}
\label{sec:datasets}
 \begin{figure*}[t]
        \resizebox{0.57\textwidth}{!}{
        \subcaptionbox{}%An example image.
        [.45\linewidth]%
        {\includegraphics[scale=0.53]{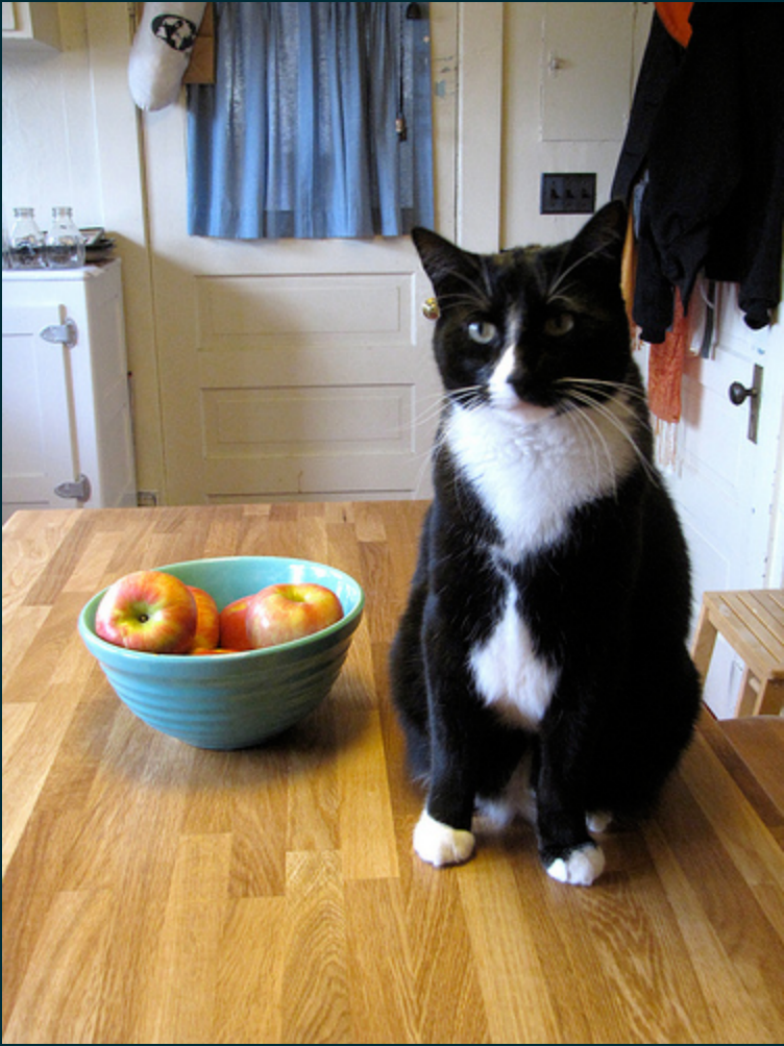}}
        \subcaptionbox{}
        [.55\linewidth]%
         {
            \begin{tabular}{l|l|l}
            \toprule
            Caption Type & Template & Example\\
            \midrule
            \textbf{attribute}                  & The (obj) is/are (attr). & The bowl is \underline{teal} (\textit{white}).\\
            \textbf{attribute\_relation}       & The (attr) (subj) is/are \{pred\} the \{obj\}. & The \underline{black and white} (\textit{gray}) cat is on the table. \\
            \midrule
            \textbf{relation}                   & The (subj) is/are (pred) the (obj). & The bowl is \underline{to the left of} (\textit{to the right of}) the cat. \\
            \textbf{relation\_attribute}        & The \{attr\} (subj) is/are (pred) the \{attr\} (obj). & The jars are to the left of the white \underline{door} (\textit{table}).\\
            \midrule
            \textbf{object\_count}              & There are (n) \{obj\}. & There are \underline{two} (\textit{three}) jars.\\
            \textbf{object\_compare\_count}     & \makecell{There are (fewer/more/as many)\\ \{obj1\} than/as \{obj2\}.} & There are \underline{more} (\textit{fewer}) apples than jars.\\
            \midrule
            \textbf{verify\_object\_attribute}   & There is (no/at least one) \{obj\} that is \{attr\}. & There is \underline{no} (\textit{at least one}) table that is plastic.\\
            \textbf{verify\_object\_relation}    & \makecell{There is (no/at least one) \{subj\} \\that is \{pred\} the \{obj\}.} & \makecell{There is \underline{at least one} (\textit{no}) cat that is \\to the right of the bowl.}\\
            \midrule
            \textbf{AND\_logic\_attribute}      & There is/are both (attr1) \{obj1\} and (attr2) \{obj2\}. & There are both a \underline{white} (\textit{metal}) door and a teal bowl.\\
            \textbf{AND\_logic\_relation}       & \makecell{There are both (subj1) (pred1) the\\ (obj1) and (subj2) (pred2) the (obj2).} & \makecell{There are both apples in the bowl and \underline{jars} (\textit{coats})\\ to the left of the door.} \\
            \textbf{XOR\_logic\_attribute}      & There is/are either (attr1) \{obj1\} or (attr2) \{obj2\}. & There is either a white door or a \underline{brown} (\textit{teal}) bowl.\\
            \textbf{XOR\_logic\_relation}       & \makecell{The \{subj\} is/are \{pred\} either \\the (obj1) or the (obj2).} & \makecell{The cat is in front of either the door\\ or the \underline{apples} (\textit{curtain}).}\\
            \bottomrule
            \end{tabular}
            }
      }
      \caption{\textbf{(a)} an illustration of one image and \textbf{(b)} exemplary captions based on the displayed caption type templates.}
      \label{fig:hnc_examples}
\end{figure*}

We use the structural information provided by scene graphs \citep{DBLP:journals/ijcv/KrishnaZGJHKCKL17} to automatically generate \textbf{hard negative image--text pairs} with various caption types. 
We leverage the ground-truth scene graphs provided by the GQA \cite{gqa} dataset, which contains a total of $+80$K~images paired with scene graphs in the training and validation set.

We define a positive caption as a textual description that \textbf{locally describes} an image, i.e.,~the caption describes a part of the image and does not aim to provide an exhaustive description of the entire scene. A hard negative caption, in turn, is \textbf{minimally contradictory} to the image and is obtained by altering a piece of information in the corresponding positive caption, i.e., without that minimal change, it would be a positive caption.

\subsection{Automatic Caption Generation}
Given an image, we first extract structured information from its corresponding scene graph and use it to create caption pairs for each of the caption types which can be found in Figure \ref{fig:hnc_examples}. 
In the caption generation process, we apply the following procedure: \textbf{1)} Check whether the information allows constructing the particular caption type. 
If yes, \textbf{2)} instantiate a positive caption with the pre-defined caption template. 
\textbf{3)} Instantiate a negative caption using the same template by replacing a piece of information in the positive caption. 
We provide an illustration of our workflow in Figure \ref{fig:workflow}.

\paragraph{Ambiguity (I)}
We apply a set of heuristics that filter out potentially ambiguous captions (see \ref{subsec:cpt-gen-setting} for details). These heuristics prevent generating captions that refer to: \textbf{a)} multiple instances of the same object class, e.g.,~\textit{the sheep that is to the right of the sheep}; \textbf{b)} relations between body parts, e.g.,~\textit{the ear is to the left of the nose}; \textbf{c)} relations between objects with one of them  typically covering a large area in the scene, e.g.,~\textit{the grass is to the left of the ball}. Note that these heuristics are applied to both the positive and the negative captions.

\paragraph{Plausible negative value sampling (II)}\label{pa:plausible_sampling} 
There are several ways to sample a negative value as the \emph{foiled} piece of information. We introduce the setting used in our experiments in the following and discuss the other options in \ref{subsec:cpt-gen-setting}. An ideal \emph{foiled} hard negative caption is \emph{visually challenging}, \emph{sensible}, and \emph{semantically similar} to the positive caption. 
To ensure that the negative caption is visually challenging, we sample a negative value from within the scene, i.e.,~the candidate values are extracted from the same scene graph. Ensuring that the negative caption is sensible and at the same time semantically similar to the positive one is more challenging. For this, we need to satisfy two conditions: 
\textbf{a)} A negative value must be valid in terms of 
semantic class constraints,~i.e., we cannot replace \underline{apple} by \underline{table} 
in \emph{The girl is eating an \underline{apple}}. 
\textbf{b)} Concept co-occurrence distributions in the negative and the positive captions should be similar to avoid spurious correlations. 
To achieve sensibility, we create look-up tables that help us define which candidates are valid for a given word. 
We then sample a negative value from these valid candidates following the distribution of the positive captions. The candidates are further filtered to avoid potential noisy replacements which we discuss in the following.

\paragraph{Noisy negative values (III)}\label{pa:noisy_negative_values}
To minimize potential issues caused by \textbf{partial} or \textbf{incomplete} scene graphs \cite{Chang2023SceneGraphSurvey}, we employ a set of heuristics designed to detect missing spatial relations between a pair of objects in a scene. 
We achieve this by leveraging the bounding-box values of the objects obtained from the ground truth scene graphs. Given a spatial relation between two entities annotated in a ground-truth scene graph; when replacing an entity or the relation with another value to create the negative caption, if this relation between the entities is not encoded in the scene graph, we check the bounding-box annotations to see if there does exist this spatial relation between the entities. If this is the case, we remove the value from the set of valid candidates\footnote{Details are given in \ref{subsec:cpt-gen-setting}.}.

\subsection{Caption Types} 
We design $12$ caption types grouped into $5$ categories, illustrated in Figure \ref{fig:hnc_examples} (together with the construction templates, an image, and examples): 
\textbf{1)} \textbf{attribute}-based, \textbf{2)} \textbf{relation}-based, \textbf{3)} \textbf{counting}-based, \textbf{4)} \textbf{existence}-based, and \textbf{5)} \textbf{reasoning}-based. 
The first three of these categories focus on either an object, an attribute, or a relation, while the existence and the reasoning-based types are some combinations of all other types. 

\paragraph{Attribute-based}
For attribute-based modality mismatches, we design two templates: \textbf{(a) attribute}, \textbf{(b) attribute\_relation}. The former simply requires models to verify whether the attribute of an object is described correctly in the caption, while the latter further challenges models' understanding of an object's attribute in a relational subgraph.

\paragraph{Relation-based}
These caption types are designed to detect a modality mismatch in relational subgraphs by foiling either the subject, the object, or the predicate to create the negative caption. 
There are two template types: \textbf{(a) relation}, \textbf{(b) relation\_attribute}. The first one aims to harness a model's sensitivity towards modality mismatches occurring in a relational subgraph. 
The second type extends the previous one by adding (an) attribute(s) to the entities in the relational subgraph, which requires a model to reason compositionally. 

\paragraph{Counting-based}
Two templates target counting-based modality mismatches:
\textbf{(a)}~\textbf{object\_count} which refers to the number of objects of the same class in the visual modality, and \textbf{(b)}~\textbf{object\_compare\_count} which compares the counts of two object classes using comparative quantifiers,~i.e., \textsl{fewer, more, as many as}, without mentioning the actual counts.

\paragraph{Existence-based}
This type addresses the existence of an entity in the visual modality.
Two templates are provided for this: \textbf{(a) verify\_object\_attribute} grounds the entity in the scene with the help of an adjective modifier, and \textbf{(b) verify\_object\_relation} does so with the help of its relation to another object in the scene.

\paragraph{Reasoning-based} 
For our reasoning-based captions, we focus on
the \textbf{AND} and \textbf{XOR} logic reasoning types. 
For each type we provide two templates, one introduces a foiled attribute and the other introduces a foil in the relational subgraph. 
These hard negative captions are very complex, and the captions contain a lot of information of which only a small piece is incorrect. 
Thus, any shortcut in reasoning should result in an incorrect prediction.

\subsection{Dataset Statistics}
We follow the official splits of the \gls{gqa} dataset \cite{gqa} to generate captions. 
The training set contains $74,942$ images, the validation set $10,696$ images.

The statistics of the \emph{clean-strict} variation of our dataset (the debiased one according to our iterative quality control explained in Section~\ref{subsec:main-cpt-generation}) is  as follows:  
For the training set we create $242$~captions for each image on average, and for the validation set $239$~captions on average, resulting in a total of $16,416,392$~for the training set and $2,314,832$~for the validation set.
The average caption length is $10$~tokens.
Due to our automatic caption generation procedure, we receive equal data distributions and caption lengths for the training and validation splits. 
Details are given in Table~\ref{tab:Splits_Full}.
\begin{table*}[t]
    \centering
    \scriptsize
    \resizebox{.9\textwidth}{!}{
    \begin{tabular}{@{~}l@{~}cccc|cccc}
    \toprule
    {} & \multicolumn{4}{c|}{\vilbert} & \multicolumn{4}{c}{\visualbert} \\
    \midrule
    {} & \volta & \foil & \hncS & \hncF & \volta & \foil & \hncS & \hncF \\
    \midrule
    attribute          & 44.1 & 52.5 & 57.5 & \textbf{78.2} & 45.0 & 51.0 & 65.7 & 77.7 \\
    attribute\_rel     & 47.5 & 54.0 & 54.3 & 75.0 & 47.5 & 49.5 & 60.4 & \textbf{79.0} \\
    relation           & 46.2 & 54.7 & 55.0 & 62.8 & 47.3 & 52.2 & 56.1 & \textbf{65.7} \\
    relation\_attr     & 47.0 & 54.4 & 55.4 & 66.4 & 47.0 & 52.8 & 61.0 & \textbf{67.0} \\
    obj\_count         & 51.0 & 49.5 & 55.9 & \textbf{73.0} & 49.0 & 48.0 & 62.7 & 66.0 \\
    obj\_comp\_count   & 50.0 & 48.5 & 57.2 & 58.5 & 48.5 & 51.0 & 58.2 & \textbf{62.0} \\
    verify\_obj\_attr  & 49.0 & 50.5 & 52.1 & \textbf{76.0} & 49.0 & 50.0 & 57.6 & 75.0 \\
    verify\_obj\_rel   & 49.5 & 51.0 & 56.3 & 59.0 & 48.5 & 48.5 & 56.5 & \textbf{61.5} \\
    AND\_logic\_attr   & 48.5 & 51.5 & 52.2 & 73.5 & 50.0 & 51.0 & 56.6 & \textbf{74.0} \\
    AND\_logic\_rel    & 52.5 & 52.0 & 52.7 & 57.0 & 48.5 & 52.0 & 52.7 & \textbf{58.5} \\
    XOR\_logic\_attr   & 50.0 & 51.0 & 51.7 & 65.5 & 52.5 & 50.0 & 57.3 & \textbf{68.0} \\
    XOR\_logic\_rel    & 51.0 & 49.5 & 57.6 & 59.0 & 51.5 & 50.5 & 57.9 & \textbf{66.5} \\
    \textbf{all}       & 48.3 & 51.6 & 54.1 & 66.4 & 48.3 & 50.5 & 58.6 & \textbf{67.9} \\ 
    \bottomrule
    \end{tabular}
    }
\caption{\protect Binary classification accuracy on \gls{hnc} test set.}
\label{tab:results-hnc-test}
\end{table*}

\section{Human-annotated Challenge Set}
\label{sec:annotation}
As we rely on scene graphs and an automatic generation procedure to create our training and validation data, we believe in the importance of providing a quality test set ideally free from any noise introduced by our automatic procedure. 
To this end, we had $19$~annotators\footnote{All students of an international \mbox{(under-)graduate} program with advanced English proficiency. We informed the participants about the use of their data and compensated them  with~$13$\texteuro/hour, above the German minimum wage.} to write down pairs of captions for all caption types.

\paragraph{Annotation guidelines}
For each image, the annotators were asked to provide a positive and a negative caption pair per their assigned caption type(s). 
We set the following conditions for the annotation: 
\textbf{1)} Stay true to the vocabulary: The words in the captions must come from within the global \gls{gqa} vocabulary. 
\textbf{2)} Choose visually challenging objects: The objects introduced as the \emph{foiled} information in the captions must come from within the scene. 
\textbf{3)} Chose linguistically challenging attributes and predicates: The attributes and predicates introduced as the \emph{foiled} information in the negative captions must be linguistically challenging, e.g.,~{brown dog → \emph{black} dog}; meaning that both captions are equally plausible. 
The annotators were instructed to skip creating a caption pair for the respective type in cases where at least one of the negative or positive captions cannot be created for a given image.

\paragraph{Dataset statistics} In total, we obtain captions for $100$~images. 
With $12$~caption types, annotation results in $3201$~captions with an average length of $8.42$.
Per caption type, we get $32$ captions on average. 
The annotated captions went under a quality check performed by another group that did not take part in the annotation.

\section{Experiments}
\label{sec:experiments}
\begin{table*}[t]
    \centering
    \scriptsize
    \resizebox{.9\linewidth}{!}{
    \begin{tabular}{@{~}l@{~}cccc|cccc}
    \toprule
    {} & \multicolumn{4}{c|}{\vilbert} & \multicolumn{4}{c}{\visualbert} \\
    \midrule
    {} & \volta & \foil & \hncS & \hncF & \volta & \foil & \hncS & \hncF \\
    \midrule
    existence               & 47.8 & 49.8 & 52.1 & 59.8 & 46.9 & 49.3 & 58.9 & \textbf{63.1} \\
    plurals                 & 50.0 & 50.4 & 51.4 & 51.4 & 49.5 & 50.3 & 51.8 & \textbf{52.8}  \\
    counting\_small\_quant  & 49.4 & 49.3 & 51.1 & 58.6 & 49.6 & 50.0 & 53.2 & \textbf{58.8}  \\
    counting\_adversarial   & 49.5 & 52.5 & \textbf{54.6} & 53.2 & 48.9 & 50.7 & 50.4 & 50.2  \\
    counting\_hard          & 49.8 & 49.6 & 49.9 & 52.4 & 49.6 & 49.7 & 50.3 & \textbf{53.2} \\
    relations               & 49.8 & 49.8 & 50.9 & 50.9 & 49.8 & 50.0 & 50.4 & \textbf{51.4} \\
    actant\_swap            & 48.1 & 54.6 & 55.8 & 58.0 & 47.9 & 51.5 & \textbf{58.3} & 57.6 \\
    action\_replacement     & 47.0 & 53.0 & 51.6 & 52.9 & 47.8 & 50.3 & 51.0 & \textbf{54.3} \\
    coreference\_standard   & 49.9 & \textbf{50.1} & 50.0 & 47.2 & 50.0 & 49.9 & 49.7 & 49.9 \\
    coreference\_hard       & \textbf{50.0} & \textbf{50.0} & \textbf{50.0} & 48.2 & \textbf{50.0} & \textbf{50.0} & 49.8 & 48.9 \\
    foil\_it                & 46.0 & \textcolor{red}{77.0} & 50.4 & 51.8 & 43.7 & \textcolor{red}{79.0} & 51.5 & \textbf{54.8} \\
    \textbf{all}            & 48.8 & 50.9 & 51.6 & 53.0 & 48.4 & 50.2 & 52.3 & \textbf{54.4} \\
    \bottomrule
    \end{tabular}
    }
\caption{\protect Binary classification accuracy on VALSE \cite{parcalabescu-etal-2022-valse} under zero-shot evaluation. For the models trained on \foil dataset, we do not calculate the accuracies obtained from the foil it splits (marked \textcolor{red}{red}) into the averaged values.}
\label{tab:valse_results}
\end{table*}

We use the \gls{volta} framework \cite{volta} as a unified testing suite to run our experiments. 
Specifically, we use its controlled setup\footnote{The controlled setup uses the same pre-training objectives and datasets across models to allow systematic comparison.} and initialize all five models from the pre-trained weights provided by \gls{volta}. 
We then further train the \gls{itm} head on the training set of both \gls{hnc} and \foil.
For a fair comparison with \foil, which is substantially smaller ($197$k data points in the training split); in addition to the full-data setting (\hncF), we include an \hncS setting subsampled to $197$k data points.
We experiment with both single-stream and dual-stream architectures and analyze their performance difference (if any): \uniter, \visualbert, \vilbert, \lxmert, \vlbert \cite{tan2019lxmert, chen2020uniter, lu2019vilbert, Li2019VisualBERTAS, Su2020VL-BERT}\footnote{Model and hyperparameter details are given in \ref{subsec:model-details}.}. To test whether training on \gls{hnc} yields similar results on more recent and bigger models, we include experiments with BLIP \cite{li2022blip}, which are presented in \ref{subsec:blip}.

\paragraph{Evaluation} 
We compare the performances of the models before and after further pre-training on \gls{hnc} on two types of tasks:
(1) Linguistic comprehension tasks, and (2) Real-world downstream reasoning tasks (Sec.~\ref{ssec:lingcomprehensiontask} and \ref{ssec:reasoningtask}, resp.).
The \hncS results are averaged over five randomly sub-sampled splits, while the rest of the results come from a single run.

\subsection{Visio-Linguistic Comprehension Tasks}
\label{ssec:lingcomprehensiontask}
\paragraph{\gls{hnc}} 
We use the manually created, high-quality test set to assess the ability of fine-grained image--text understanding (see Sec. \ref{sec:datasets} for details about the automatically-created training and validation sets and Sec. \ref{sec:annotation} for the human-annotated test set).

\paragraph{\gls{valse}} is a benchmark focusing on various linguistic phenomena \cite{parcalabescu-etal-2022-valse}.

\subsection{Real-World Reasoning Tasks}
\label{ssec:reasoningtask}
\paragraph{\gls{cpt}} measures the commonsense knowledge level of task-agnostic visually pre-trained models on the  \cwwv dataset \cite{yang-silberer-2022-visual}. 
We consider this task as a real-world scenario in that associated images are automatically retrieved, which may lead to noisy image--text pairs (see \ref{pa:cpt_analysis} for the complete task description).

\paragraph{\gls{gqa}} is a dataset designed for real-world visual reasoning and compositional question answering. 
Unlike the aforementioned tasks that test zero-shot capabilities, we investigate whether our weight initialization after \gls{hnc} further pre-training serves as an improved starting point when fine-tuning on \gls{gqa}. 
Therefore, we compare \volta checkpoints and further pre-trained ones (\hnc) after their fine-tuning on \gls{gqa}.
The performances are reported on the \gls{gqa} testdev split.

\section{Results}
\label{sec:results}
We report the results, i.e.~classification accuracies, on the aforementioned four tasks\footnote{We only discuss the statistically significant results.}.
We compare dual-stream and single-stream models to assess the effects of different modality integration methods on models' ability to detect mismatches. 
We display the results obtained from our further pre-trained weight initializations as \hncS and \hncF, the ones obtained from training on FOIL-COCO as \foil, and the official \gls{volta} weight initialization as \volta\footnote{We only display results from one single- and one dual-stream model in Table \ref{tab:results-hnc-test}, \ref{tab:valse_results}, \ref{tab:results-cwwv}, and \ref{tab:gqa_results}. Complete results can be found in Table \ref{tab:results-hnc-complete}, \ref{tab:results-valse-complete}, \ref{tab:results-cwwv-complete}, and \ref{tab:results-gqa-complete} resp. in \ref{sec:appendix}.}. 
The best results are shown in \textbf{bold}.

\begin{table*}[t]
    \centering
    \scriptsize
    \resizebox{0.9\textwidth}{!}{
    \begin{tabular}{lcccc|cccc}
    \toprule
     & \multicolumn{4}{c}{\lxmert} & \multicolumn{4}{c}{\uniter} \\
    \midrule
     & \volta & \foil & \hncS & \hncF & \volta & \foil & \hncS & \hncF \\
    \midrule
    taxonomic &  51.55& 52.46& 54.78& 54.8 & 54.04& 56.69& 57.29& \textbf{58.5} \\
    similarity & 43.01& 43.17& 44.38& 46.43& 46.43& 49.53& 50.99& \textbf{55.75} \\
    part-whole & 53.73& 50.13& 52.93& 56.48& 63& 63.95& 64.1& \textbf{69.01} \\
    spatial & 55.6& 52.72& 55.04& 56.79& 57.41& 57.47& 53.32& \textbf{57.97} \\
    temporal & 49.23& 49.81& 47.56& \textbf{50.24}& 47.86& 46.59& 46.53& 46.27 \\
    all & 55.43& 52.22& 55.94& 55.49& 58.53& 59.29& 59.27& \textbf{62.32} \\
    \bottomrule
    \end{tabular}
    }
\caption{\protect Classification accuracy on \gls{cpt} \cite{yang-silberer-2022-visual} with \cwwv under zero-shot evaluation.} 
\label{tab:results-cwwv}
\end{table*}

\subsection{Visio-Linguistic Comprehension Tasks}
\paragraph{\gls{hnc}}
Table \ref{tab:results-hnc-test} displays the results obtained on our human-annotated test set. 
Zero-shot performances of \gls{volta} checkpoints on the majority of the caption types are close to random baseline ($50\%$) showing that the dataset is not trivially solvable. 
We observe a strong \textbf{under-prediction of entailment\footnote{False negative prediction for the positive pairs.}} in models initialized from \gls{volta} checkpoints before undergoing our further pre-training on \gls{hnc} dataset, suggesting that the positive captions are equally hard to align with the visual modality for these models. 
This might be because the web-retrieved captions lack compositionally complex information, i.e.,~information about multiple objects along with their attributes or relations to other objects. 
After further pre-training on \gls{hnc} (see Tab.\ref{tab:results-hnc-test}, col.\hncF), we observe a large improvement in all caption types which showcases the effectiveness of our dataset in teaching fine-grained alignment of the visual and textual modality.

\paragraph{\gls{valse}}
As shown in Table \ref{tab:valse_results}, further pre-training on \gls{hnc} largely improves: \textbf{existence}, \textbf{counting\_small\_quant}, \textbf{counting\_adversarial}, \textbf{counting\_hard}, \textbf{actant\_swap}, \textbf{action\_replacement}, and \textbf{foil\_it}\footnote{We provide more findings with analysis in \ref{pa:valse_analysis}.}. Also, \hncS achieves better results compared to \foil on average, which suggests that \gls{hnc} contains more diverse and better quality captions to learn from than FOIL-COCO.
The large improvement we observe in \textbf{existence} type in \gls{valse} shows the effectiveness of our existence-based captions (verify\_obj\_attr, verify\_obj\_rel). 
We attribute the large improvement in \textbf{actant\_swap} to our dedicated control of subjects and objects in relational captions (relation\_subj, relation\_obj, AND\_logic\_rel, and XOR\_logic\_rel). As for the \textbf{foil\_it}, we see a similar effect, i.e.,~controlling nouns (subjects and objects) in hard negatives helps models to better ground the object in the visual scene and not be confused by another (potentially semantically similar) object.

\textbf{Counting\_adversarial} tests for the shortcut biases by purposefully assigning a more common number as the \emph{foiled} information in the \gls{valse} captions where the original caption contains a number that is typically less common in these models' pre-training data. 
Not only do we see a large performance increase in \textbf{counting\_small\_quant}, we also see an improvement in \textbf{counting\_adversarial} and \textbf{counting\_hard} captions showing that the models benefit from the diverse number sampling in \gls{hnc}'s training data construction. 

Further, we only observe a marginal improvement in \textbf{plurality} which is not surprising as we do not create captions that target this type specifically. 
Also, \gls{hnc} pre-training does not affect \textbf{coreference\_standard} and \textbf{coreference\_hard} too much (slight performance decrease if any). 
Just like the \textbf{plurality}, we expect these numbers as we do not address such types in this work. 
Future work can easily extend to \textbf{plurality} by creating a caption type that solely controls the information on the plurality of the objects in the scene. 
The same can be done for \textbf{coreference} by combining several pieces of information about an entity using a referent word.

\subsection{Real-World Reasoning Tasks}
\paragraph{\gls{cpt}} Table \ref{tab:results-cwwv} shows substantial zero-shot performance gains after further pre-training on \hncF; particularly on single-stream models. We speculate that our \gls{hnc} pre-training could drive single-stream encoders to be more sensitive towards cross-modal inconsistencies and strengthen the importance of the textual modality under noisy visual input scenarios. For dual-stream models, the overall improvement is limited, possibly due to the design of certain layers that primarily perform inter-modal attention 
which restricts the flexibility of balancing the influence of different modality inputs during inference. Regarding the individual commonsense dimensions, all \gls{hnc} models demonstrate improvement on \textbf{taxonomic}, \textbf{similarity}, \textbf{part-whole}. This could be explained by their sparser distribution of concrete concepts \cite{yang-silberer-2022-visual}, resulting in less semantic correspondence between the extracted images and their textual counterparts (see \ref{pa:cpt_analysis}, Fig.\ref{fig:cpt-examples-hnc-correct}). Overall, the outcome suggests the importance of having hard negative captions in \gls{itm} pre-training to enhance the robustness of \gls{vl} models in handling noisy visual inputs during inference. Both the scale and the quality play a role, as models show greater improvement on these dimensions when further pre-trained on \hncS compared to FOIL-COCO (see col.\foil \& \hncS of tab.\ref{tab:results-cwwv}).
However, the hard negative pretraining does not benefit much to \textbf{spatial} and \textbf{temporal}.
Especially for \textbf{temporal}, the question token and the image retrieved for the answer token are subject to mismatches due to the natural temporal order, e.g.,~\textit{run out of money} is a consequence of \textit{buying food}, the image of \textit{money} does not correspond to \textit{food} (see \ref{pa:cpt_analysis},
Fig.\ref{fig:cpt-examples-hnc-dual-incorrect-temporal}). 

\paragraph{\gls{gqa}}
We summarize our results on the \gls{gqa} \cite{gqa} testdev split in Table \ref{tab:gqa_results}.
As we are required to fine-tune on \gls{gqa} to receive meaningful results, we distinguish between the weight initialization from the official \gls{volta} pre-training and the initialization from our further pre-training on \gls{hnc}.
At first glance, our initialization points achieve higher accuracy across all five models.
The results are statistically significant for \lxmert, \uniter, and \visualbert.
For the single-stream models, \visualbert benefits the most from further pre-training on \gls{hnc}.
For the dual-stream, \lxmert shows larger performance gains.
Generally, the dual-stream vs. single-stream modality integration does not seem to have an influence on how much the respective models benefit from further pre-training on \gls{hnc}.
Nonetheless, the overall results support our hypothesis that further pre-training \gls{vl} models on more fine-grained mismatching data (in the form of hard-negative captions) improves models' cross-modal reasoning capabilities.

\begin{table}[ht]
    \centering
    \scriptsize
    \resizebox{0.49\textwidth}{!}{
    \begin{tabular}{lcc|cc}
    \toprule
    {} & \multicolumn{2}{c|}{\lxmert} & \multicolumn{2}{c}{\visualbert}  \\
    \midrule
    {} & \volta & \hncF & \volta & \hncF  \\
    \midrule
    Accuracy   & 53.48 & 55.45 & 53.51 & \textbf{56.85} \\
    \bottomrule
    \end{tabular}
    }
\caption{
Results on the \gls{gqa} \cite{gqa} testdev split.}
\label{tab:gqa_results}
\end{table}

\section{Dataset Analysis}
\label{sec:analysis}
Next, we analyze our caption generation process: how robust are the different negative sampling strategies, and which results in less/more linguistic bias that a model could exploit as a shortcut? 
We discuss the challenges of automatic hard negative caption generation, the biases introduced in captions as a result of this automatic procedure, and how to mitigate them.
We then perform a modality ablation study to ensure the quality of our human-annotated test set. We provide further qualitative analyses in Appendix \ref{subsec:qualitative-analysis}.

\subsection{Caption Generation: An Iterative Process}
\label{subsec:main-cpt-generation}
Our final caption generation process is a product of a series of refinement iterations. 
At each iteration, we train and evaluate a \gls{lm} (BERT, \citealp{devlin-etal-2019-bert}) on our captions and use the accuracy scores as a proxy to measure linguistic bias. 
Throughout this process, we found that, for example, replacing an attribute of a visual object with another attribute from the scene without any further constraint introduces a strong linguistic bias, e.g., \emph{a purple dog} (see \ref{pa:dataset-gen-sampling}). Similarly, for example, replacing an object in a (subject, predicate, object) triple by another \emph{similar} or a \emph{probable} one is rather challenging. Depending on the heuristics employed to determine what might be a \emph{probable} replacement, the resulting negative captions contain more or less linguistic bias (\gls{lm} acc. of approx. $58\%$ for \textsl{strict} constraints and approx. $66\%$ when these constraints are \textsl{relaxed}) Moreover, we discovered that the relations in scene graphs are rather sparse which, if not handled correctly, results in noisy negative captions, i.e.,~the negative caption does not contradict the image.
We provide further detailed analyses along with examples in Appendix \ref{subsec:dataset-gen}.

\subsection{Sanity Check with Modality Ablation}
\label{subsec:main-modality-ablation}
We evaluate the \gls{hnc} models under the \emph{blind} setting\footnote{The image features are $0$-masked during inference.} (see  \ref{subsec:blind-details} for details on the implementation). 
Our findings\footnote{Further analyses are provided in \ref{subsec:modality-ablation}.} suggest that the effect of world priors, especially for object quantities, is difficult to overcome in negative caption generation.\footnote{Blind VL models achieve $+3pp$.~on average in \textbf{object\_count} (Tab.~\ref{tab:hnc-test-blind} and in col. \emph{Clean Strict} in Tab.~\ref{tab:lm-test-set}).}
For example, a typical quantity of a sofa in a living room is \textit{one}. 
A negative caption with a different count of sofa violates the worldviews of \gls{vl} models. 
\gls{vlms}, being trained on typical real-world scenes, usually do not capture other counts of sofas, and as a consequence, corresponding negative captions are easier to be detected as a mismatch, even though the model is not exposed to the visual input during inference. This poses a major challenge to \gls{vl} pre-training in terms of learning modality mismatches.

\section{Conclusion}
\label{sec:conclusion}
In this work, we introduced Hard Negative Captions~(\gls{hnc}), a dataset for further pre-training Vision and Language Models to improve their modality integration capabilities on a fine-grained level and demonstrated improvements across models and tasks.
We proposed a novel automatic dataset construction procedure for constructing hard negative captions to be used for Image-Text-Matching (\gls{itm}) training as well as a challenging test set annotated by humans. We provided detailed analyses of the challenges in automatic creation of hard negative captions and proposed methods to mitigate them.
Lastly, we demonstrated the benefits of \gls{hnc} by obtaining significant model performance gains on various tasks, including the diagnostic dataset \gls{valse}, our \gls{hnc} test set as well as a commonsense probing task (\gls{cpt}), and down-stream performance gains after supervised fine-tuning on \gls{gqa}, both of which require real-world reasoning.

\section{Limitations}
\label{sec:limitations}
Automatic caption generation has its limitations. 
First, since our generation pipeline is seeded with the scene graphs \citep{DBLP:journals/ijcv/KrishnaZGJHKCKL17}, issues identified in the literature like a skewed distribution of predicates \cite{ijcai2020p82}, limited vocabulary size \cite{DBLP:conf/eccv/HeGSL22}, low-level annotations, and reference ambiguity \cite{DBLP:journals/corr/abs-2106-08543} might persist in our generated captions. 
Although we showed that certain biases can be mitigated (or minimized), our quantitative and qualitative analyses suggest that automatically generated captions based on scene graphs are subject to linguistic and distributional biases which are difficult to combat. 
Therefore, we believe that our hard negative caption generation could benefit from existing scene graph debiasing methods \cite{DBLP:conf/mm/ChiouDYWZF21}. 
Also, our method of eliminating noisy captions caused by sparse scene graph annotations is based on rule-based heuristics. 
Although it helps us avoid creating false negative captions, it does not address the issue of annotation sparseness in scene graphs. 
For a potentially more robust method, the integration of an object detector \cite{obj_detection} can be studied in future work. 
Moreover, our rule-based heuristics are specific to our use case, and they might not work for other scenarios. Nevertheless, our framework allows for easy adaptation or extension to cover a wide range of domains and tasks. Last, our contribution is mainly on the creation of training and test data for \gls{itm}. We have not investigated the impacts of our data in combination with other training objectives or methods. We leave this (and the previous points) to future work.

\section{Acknowledgement}
Funded by Deutsche Forschungsgemeinschaft (DFG, German Research Foundation) under Germany's Excellence Strategy - EXC 2075 – 390740016. We acknowledge the support by the Stuttgart Center for Simulation Science (SimTech).

\bibliography{anthology,custom}

\bibliographystyle{acl_natbib}

\clearpage

\appendix

\section{Appendix}
\label{sec:appendix}

\subsection{Model Details}
\label{subsec:model-details}

\subsubsection{\gls{itm} Objective}
\label{subsec:itm-details}
Both single and dual-stream models aim to learn an alignment between the visual and textual modality to infer the correct entailment between them. Image--text matching is the objective of inferring a similarity score between these modalities. As such, in \gls{vl} Transformers \cite{Vaswani2017Attention}, it is implemented in the form of a binary classification head that learns to predict whether an image and a text entail one another. 

\subsubsection{BLIP}\label{subsec:blip}
\gls{blip} \cite{li2022blip} is a \gls{vl} pre-training framework which is designed to perform both \gls{vl} generation and understanding tasks.
\citet{li2022blip} propose three versions of \gls{blip}: trained to align vision and language representations using an image-text contrastive loss, vision and language interactions using \gls{itm}, and a \gls{lm} loss to generate captions.
In the following, we refer to the \gls{blip} version trained with a \gls{itm} loss as \textit{\gls{blip}-\gls{itm}}.
In our experiments, we evaluated and fine-tuned \gls{blip}-\gls{itm}, since it matches the design of our \gls{hnc} dataset that aims for teaching the model's a detailed understanding of the visual input using carefully sampled negative captions.

\subsubsection{\gls{blip} Hyperparameters}
We use AdamW \cite{loshchilov2017decoupled} with a learning rate of $1e-5$ and a weight-decay of $0.05$ as used by \cite{li2022blip} to train \gls{blip}.
To fine-tune the model, we initialize a learning rate scheduler with a warm-up duration of four epochs and a starting learning of $1e-7$.
Afterward, the learning rate decays by a factor of $\gamma=0.85$.
We perform early stopping on the validation set, and train for a maximum of $20$ epochs.
The batch size during training equals $50$, and we use eight NVIDIA A100 GPUs with $80$GB VRAM.

\subsubsection{\gls{volta} Hyperparameters}
\label{subsec:hyperparameters}
\paragraph{Further pre-training on \gls{hnc}} The following hyperparameters for the \gls{volta} models are used:
ADAM optimizer \cite{adam} with a learning rate and weight decay of $4e-5$, $\beta=(0.9, 0.999)$, and gradient clipping \cite{grad_clip} with a norm of $1.0$. For the tokenizer, we used a maximum sequence length of $40$. The maximum number of regions is set to 36 just like the \gls{volta} implementations. For the training, we used a batch size of 1024 and a maximum number of epochs of 20 with early stopping. We left all other hyperparameters untouched (e.g.,~model hyperparameters), and stick with the ones provided by \gls{volta}. We used 4 NVIDIA RTX A6000 GPUs and trained the models for a maximum of $48$ hours. We use the controlled setup in \gls{volta}, which uses the same pre-training objectives and datasets across models to allow systematic comparison.

\paragraph{Fine-tuning on \gls{gqa}} For fine-tuning the \gls{volta} model checkpoints on the \gls{gqa} dataset, we use a batch size of 1024 and a maximum number of epochs of $20$ with early stopping. 
The maximum sequence length and the maximum number of regions were kept the same as in the pre-training. 
The rest of the hyperparameters are: ADAM optimizer \cite{adam} with a learning rate and weight decay of $4e-5$, $\beta=(0.9, 0.999)$, and gradient clipping \cite{grad_clip} with a norm of $5.0$. We used 2 NVIDIA RTX A6000 GPUs and trained the models for maximum $8$ hours.

For fine-tuning the \gls{blip} model checkpoints on the \gls{gqa} dataset, we use a batch size of $50$ and train for a maximum of $20$ epochs while performing early stopping.
We again use AdamW \cite{loshchilov2017decoupled} with a learning rate of $5e-5$ and weight decay of $0.05$.
The learning rate scheduler is initialized with a starting learning rate of $1e-8$, a warmup duration of three epochs, and a $\gamma=0.85$ that scales the learning rate after each epoch.

\paragraph{Language model training} We trained a BERT\footnote{bert-base-uncased} \cite{devlin-etal-2019-bert} model to predict whether a caption is positive or negative without seeing the image. The model is initialized with the pre-trained weights loaded from HuggingFace library\footnote{https://huggingface.co/}. We added a binary classification head and trained the model on \gls{hnc} captions with the entailment labels of $0$ and $1$. Following hyperparameters were used: ADAM optimizer \cite{adam} with a learning rate of $16e-5$, maximum sequence length of $40$ for the tokenizer, batch size of $8384$, maximum number of epochs $40$ with early stopping. We used a single NVIDIA RTX A6000 GPU and trained the models for maximum $120$ hours.

\subsubsection{Blind Setting in \gls{vl} Models}
\label{subsec:blind-details}
For consistency, we used the \gls{volta} implementations of the models and did not alter anything but the image features. We used $0$-masking to create the blind setting. Specifically, we create a $0$ tensor as the size of the image features and feed this into the model instead of the real image features. We do not change anything on the input of the textual modality.

\clearpage
\begin{sidewaystable}
    %\scriptsize
    \resizebox{0.99\textwidth}{!}{
    \begin{tabular}{lcccc|cccc|cccc|cccc|cccc|cc}
    \toprule
    {} & \multicolumn{8}{c|}{Dual-Stream} & \multicolumn{12}{c|}{Single-Stream} & \multicolumn{2}{c}{Contrastive Dual-Encoder}\\
    \midrule
    {} & \multicolumn{4}{c|}{\vilbert} & \multicolumn{4}{c|}{\lxmert} & \multicolumn{4}{c|}{\uniter} & \multicolumn{4}{c|}{\visualbert} & \multicolumn{4}{c|}{\vlbert} & \multicolumn{2}{c}{BLIP}\\
    \midrule
    {} & \volta & \foil & \hncS & \hncF & \volta & \foil & \hncS & \hncF & \volta & \foil & \hncS & \hncF & \volta & \foil & \hncS & \hncF & \volta & \foil & \hncS & \hncF & ITM & \hncF\\
    \midrule
    attribute          & 44.1 & 52.5 & 57.5 & \textbf{78.2} & 45.5 & 54.5 & 59.7 & 76.2 & 42.6 & 50.5 & 67.4 & \textbf{77.7} & 45.0 & 51.0 & 65.7 & \textbf{77.7} & 41.1 & 52.0 & 64.4 & 75.7 & 55.0 & \textbf{80.2} \\
    attribute\_rel     & 47.5 & 54.0 & 54.3 & \textbf{75.0} & 50.5 & 49.0 & 55.6 & 74.5 & 47.0 & 51.5 & 61.5 & 75.0 & 47.5 & 49.5 & 60.4 & \textbf{79.0} & 47.5 & 51.5 & 59.2 & 72.0 & 57.0 & \textbf{77.0} \\
    relation           & 46.2 & 54.7 & 55.0 & 62.8 & 47.7 & 53.8 & 54.7 & \textbf{64.7} & 45.8 & 53.8 & 55.2 & 63.2 & 47.3 & 52.2 & 56.1 & \textbf{65.7} & 47.0 & 53.5 & 55.2 & 63.8 & 56.0 & \textbf{64.7} \\
    relation\_attr     & 47.0 & 54.4 & 55.4 & 66.4 & 48.6 & 56.1 & 57.8 & \textbf{67.0} & 46.2 & 52.8 & 61.4 & \textbf{67.8} & 47.0 & 52.8 & 61.0 & 67.0 & 46.2 & 54.1 & 59.4 & 67.5 & 56.3 & \textbf{70.3} \\
    obj\_count         & 51.0 & 49.5 & 55.9 & \textbf{73.0} & 49.0 & 51.0 & 55.3 & 71.5 & 46.5 & 50.0 & 59.1 & \textbf{68.5} & 49.0 & 48.0 & 62.7 & 66.0 & 51.5 & 50.0 & 63.0  & 65.0 & 54.0 & \textbf{72.0} \\
    obj\_comp\_count   & 50.0 & 48.5 & 57.2 & \textbf{58.5} & 49.0 & 51.0 & 55.6 & 54.5 & 49.5 & 51.0 & 56.6 & 54.5 & 48.5 & 51.0 & 58.2 & 62.0 & 49.0 & 50.5 & 60.9 & \textbf{62.5} & 49.5 & \textbf{55.0} \\
    verify\_obj\_attr  & 49.0 & 50.5 & 52.1 & \textbf{76.0} & 48.5 & 49.5 & 49.8 & 65.5 & 50.5 & 50.0 & 61.3 & 73.0 & 49.0 & 50.0 & 57.6 & \textbf{75.0} & 50.0 & 49.5 & 54.2 & 72.0 & 51.5 & \textbf{73.5} \\
    verify\_obj\_rel   & 49.5 & 51.0 & 56.3 & 59.0 & 51.0 & 49.5 & 54.8 & \textbf{64.0} & 50.0 & 49.5 & 56.8 & 63.5 & 48.5 & 48.5 & 56.5 & 61.5 & 49.5 & 50.0 & 58.3 & \textbf{66.0} & 49.5 & \textbf{67.0} \\
    AND\_logic\_attr   & 48.5 & 51.5 & 52.2 & 73.5 & 50.0 & 48.5 & 54.4 & \textbf{70.0} & 49.5 & 50.5 & 54.4 & 71.0 & 50.0 & 51.0 & 56.6 & 74.0 & 49.0 & 50.5 & 55.4 & \textbf{74.5} & 55.0 & \textbf{77.5} \\
    AND\_logic\_rel    & 52.5 & 52.0 & 52.7 & \textbf{57.0} & 51.0 & 50.0 & 51.0 & 56.5 & 49.5 & 52.5 & 51.1 & \textbf{59.5} & 48.5 & 52.0 & 52.7 & 58.5 & 51.5 & 52.5 & 52.8 & 57.5 & 54.5 & \textbf{58.5} \\
    XOR\_logic\_attr   & 50.0 & 51.0 & 51.7 & 65.5 & 49.0 & 48.5 & 51.0 & \textbf{67.0} & 51.0 & 49.0 & 55.6 & 65.0 & 52.5 & 50.0 & 57.3 & \textbf{68.0} & 51.5 & 50.5 & 54.9 & 67.0 & 43.5 & \textbf{73.0} \\
    XOR\_logic\_rel    & 51.0 & 49.5 & 57.6 & 59.0 & 51.5 & 49.5 & 60.2 & \textbf{64.5} & 51.0 & 49.0 & 58.3 & 68.0 & 51.5 & 50.5 & 57.9 & 66.5 & 50.0 & 50.5 & 58.7 & \textbf{70.5} & 49.5 & \textbf{65.0} \\
    \textbf{all}       & 48.3 & 51.6 & 54.1 & \textbf{66.4} & 49.0 & 50.9 & 55.0 & 66.2 & 47.7 & 50.8 & 58.2 & 67.1 & 48.3 & 50.5 & 58.6 & \textbf{67.9} & 48.1 & 51.3 & 58.0 & 67.3 & 53.5 & \textbf{69.0} \\ 
    \bottomrule
    \end{tabular}
    }
    \caption{\protect Binary classification accuracy on \gls{hnc} test set. The column BLIP-ITM refers to the checkpoint of BLIP that was fine-tuned on COCO and Flickr30k for image-text retrieval.} 
    \label{tab:results-hnc-complete}
\end{sidewaystable}
\begin{sidewaystable}
    \resizebox{0.99\textwidth}{!}{
    \begin{tabular}{lcccc|cccc|cccc|cccc|cccc|cc}
    \toprule
    {} & \multicolumn{8}{c|}{Dual-Stream} & \multicolumn{12}{c|}{Single-Stream} & \multicolumn{2}{c}{Contrastive Dual-Encoder}\\
    \midrule
    {} & \multicolumn{4}{c|}{\vilbert} & \multicolumn{4}{c|}{\lxmert} & \multicolumn{4}{c|}{\uniter} & \multicolumn{4}{c|}{\visualbert} & \multicolumn{4}{c|}{\vlbert} & \multicolumn{2}{c}{BLIP}\\
    \midrule
    {} & \volta & \foil & \hncS & \hncF & \volta & \foil & \hncS & \hncF & \volta & \foil & \hncS & \hncF & \volta & \foil & \hncS & \hncF & \volta & \foil & \hncS & \hncF & ITM & \hncF\\
    \midrule
    existence               & 47.8 & 49.8 & 52.1 & 59.8 & 47.3 & 52.1 & 52.2 & \textbf{60.3} & 46.2 & 49.8 & 63.5 & \textbf{66.9} & 46.9 & 49.3 & 58.9 & 63.1 & 47.1 & 49.7 & 58.6 & 59.2 & 52.06 & \textbf{57.77} \\
    plurals                 & 50.0 & 50.4 & \textbf{51.4} & \textbf{51.4} & 49.9 & 50.1 & 50.2 & 50.3 & 50.0 & 50.1 & 51.3 & 50.7 & 49.5 & 50.3 & 51.8 & \textbf{52.8} & 49.6 & 50.4 & 51.6 & 51.3 & \textbf{57.95} & 55.60 \\
    counting\_small\_quant  & 49.4 & 49.3 & 51.1 & \textbf{58.6} & 49.5 & 49.1 & 50.4 & 55.9 & 49.4 & 49.7 & 53.2 & 57.5 & 49.6 & 50.0 & 53.2 & \textbf{58.8} & 49.5 & 49.4 & 52.5 & 58.1 & 53.30 & \textbf{63.05} \\
    counting\_adversarial   & 49.5 & 52.5 & \textbf{54.6} & 53.2 & 49.6 & 50.7 & 54.0 & 50.2 & 49.1 & 50.7 & \textbf{54.2} & 53.6 & 48.9 & 50.7 & 50.4 & 50.2 & 49.5 & 50.0 & 53.3 & 52.7 & 55.96 & \textbf{57.76} \\
    counting\_hard          & 49.8 & 49.6 & 49.9 & 52.4 & 49.9 & 50.4 & 50.5 & \textbf{52.6} & 49.9 & 50.0 & 51.0 & 52.6 & 49.6 & 49.7 & 50.3 & 53.2 & 49.7 & 49.8 & 51.5 & \textbf{53.3} & 53.46 & \textbf{58.02} \\
    relations               & 49.8 & 49.8 & 50.9 & 50.9 & 50.0 & 49.9 & 50.5 & \textbf{51.6} & 50.0 & 50.0 & 50.3 & 50.7 & 49.8 & 50.0 & 50.4 & \textbf{51.4} & 49.9 & 50.2 & 50.7 & 50.1 & \textbf{54.97} & 54.07 \\
    actant\_swap            & 48.1 & 54.6 & 55.8 & \textbf{58.0} & 49.0 & 53.2 & 55.4 & 57.4 & 47.9 & 51.0 & 54.4 & 55.9 & 47.9 & 51.5 & \textbf{58.3} & 57.6 & 48.3 & 50.4 & 53.8 & 56.3 & 52.47 & \textbf{57.41}\\
    action\_replacement     & 47.0 & 53.0 & 51.6 & 52.9 & 48.1 & 52.1 & 53.2 & \textbf{55.1} & 47.5 & 51.0 & 52.0 & 52.8 & 47.8 & 50.3 & 51.0 & \textbf{54.3} & 47.4 & 50.5 & 49.7 & 52.3 & \textbf{61.94} & 58.41\\
    coreference\_standard   & 49.9 & \textbf{50.1} & 50.0 & 47.2 & 50.0 & 50.1 & 49.8 & 48.9 & 49.9 & 50.1 & 49.8 & 50.1 & 50.0 & 49.9 & 49.7 & 49.9 & 50.0 & 49.9 & 49.9 & \textbf{50.2} & \textbf{50.66} & 50.11\\
    coreference\_hard       & \textbf{50.0} & \textbf{50.0} & \textbf{50.0} & 48.2 & \textbf{50.0} & \textbf{50.0} & 49.5 & 48.6 & 50.0 & 50.0 & 50.1 & 50.4 & 50.0 & 50.0 & 49.8 & 48.9 & 50.0 & 50.0 & 49.4 & \textbf{51.1} & 50.35 & \textbf{53.55}\\
    foil\_it                & 46.0 & \textcolor{red}{77.0} & 50.4 & 51.8 & 45.8 & \textcolor{red}{77.5} & 50.1 & 52.4 & 43.8 & \textcolor{red}{77.1} & 51.2 & 54.1 & 43.7 & \textcolor{red}{79.0} & 51.5 & \textbf{54.8} & 44.2 & \textcolor{red}{78.1} & 50.0 & 53.6 & \textbf{72.10} & 59.20 \\
    \textbf{all}            & 48.8 & 50.9 & 51.6 & \textbf{53.0} & 48.9 & 50.8 & 51.4 & \textbf{53.0} & 48.4 & 50.2 & 52.8 & 53.9 & 48.4 & 50.2 & 52.3 & \textbf{54.4} & 48.5 & 50.0 & 51.9 & 53.5 & \textbf{57.16} & 57.13 \\
    \bottomrule
    \end{tabular}
    }
    \caption{\protect Binary classification accuracy on VALSE \cite{parcalabescu-etal-2022-valse} under zero-shot evaluation. For the models trained on \foil dataset, we do not calculate the accuracies obtained from the foil it splits (marked \textcolor{red}{red}) into the averaged values. The column BLIP-ITM refers to the checkpoint of BLIP that was fine-tuned on COCO and Flickr30k for image-text retrieval.} 
    \label{tab:results-valse-complete}
\end{sidewaystable}

\begin{sidewaystable}
    %\scriptsize
    \resizebox{0.99\textwidth}{!}{
        \begin{tabular}{lcccc|cccc|cccc|cccc|cccc}
        \toprule
        {} & \multicolumn{8}{c|}{Dual-Stream} & \multicolumn{12}{c}{Single-Stream} \\
        \midrule
        {} & \multicolumn{4}{c|}{\vilbert} & \multicolumn{4}{c|}{\lxmert} & \multicolumn{4}{c|}{\uniter} & \multicolumn{4}{c|}{\visualbert} & \multicolumn{4}{c}{\vlbert} \\
        \midrule
        {} & \volta & \foil & \hncS & \hncF & \volta & \foil & \hncS & \hncF & \volta & \foil & \hncS & \hncF & \volta & \foil & \hncS & \hncF & \volta & \foil & \hncS & \hncF \\
        \midrule
            part-whole   & 55.02& 54.59& 52.65& 55.97& 53.73& 50.13& 52.93& 56.48& 63& 63.95& 64.1& \textbf{69.01}& 63.35& 64.46& 66.47& 68.84& 58.37& 61.37& 59.43& 63.43 \\
            distinctness &  55.92& 56.76& 54.56& 54.83& 59.42& 55.68& 61.98& 60.51& 65.94& 67.27& 66.42& 70.65& 65.7& 69.69& 70.89& 68.84& 66.06& \textbf{71.38}& 67.08& 68.6       \\
            similarity   & 42.24& 40.06& 43.11& 41.3& 43.01& 43.17& 44.38& 46.43& 46.43& 49.53& 50.99& \textbf{55.75}& 48.45& 50.16& 53.26& 52.02& 47.2& 49.22& 48.88& 49.69          \\
            temporal     & 47.17& 45.79& 41.5& 39.28& 49.23& 49.81& 47.56& 50.24& 47.86& 46.59& 46.53& 46.27& 50.98& 51.51& 50.19& 52.3& 51.83& 51.67& 50.5& \textbf{53.31} \\
            taxonomic    & 49.89& 52.38& 52.53& 52.08& 51.55& 52.46& 54.78& 54.8& 54.04& 56.69& 57.29& 58.5& 56.84& 61.75& 61.34& \textbf{63.27}& 55.4& 62.21& 57.93& 62.74 \\
            quality       & 57.39& 56.63& 55.64& 56.85& 62.45& 64.18& 67.2& \textbf{70.11}& 61.58& 54.57& 62.8& 62.17& 61.2& 65.6& 68.17& 66.9& 62.23& 63.48& 65.52& 68.04 \\
            spatial      & 52.91& 51.97& 45.41& 50.47& 55.6& 52.72& 55.04& 56.79& 57.41& 57.47& 53.32& 57.97& 58.1& \textbf{58.54}& 57.01& 57.54& 56.47& 56.66& 53.85& 57.47 \\
            utility     & 60.48& 57.8& 57.57& 57.32& 62.87& 57.51& 63.91& 60.62& 65.36& 64.93& 65.95& 68.71& 67.22& 68.66& 67.79& \textbf{70.33}& 65.12& 65.74& 65.15& 65.74 \\
            desire       & 56.6& 54.41& 53.23& 47.05& 54.8& 51.88& 54.6& 52.05& 58.51& 57.72& 58.75& \textbf{59.63}& 57.27& 57.78& 58.34& 57.22& 50.42& 52.22& 50.48& 51.43 \\
            creation &  52.0& 54.0& 58.6& 53.0& 56.0& 54.0& 53.4& 54.0& 62& 64& 63.2& \textbf{77}& 65& 66& 68.4& 70& 63& 66& 64.4& 67 \\
            \textbf{all }    & 53.95& 52.99& 51.15& 50.87& 55.43& 52.22& 55.94& 55.49& 58.53& 59.29& 59.27& \textbf{62.32}& 59.24& 61.2& 61.5& 62.16& 57.42& 59.33& 57.88& 60.28 \\
        \bottomrule
        \end{tabular}
    }
    \caption{\protect  Classification accuracy on \gls{cpt} \cite{yang-silberer-2022-visual} task under zero-shot evaluation. 
    } 
    \label{tab:results-cwwv-complete}
\end{sidewaystable}
\twocolumn

\begin{table*}[ht]
    \centering
    \scriptsize
    \resizebox{1\textwidth}{!}{
    \begin{tabular}{@{~}l@{~}rr|rr|rr|rr|rr|rrr}
    \toprule
    %{} & \multicolumn{4}{c|}{Dual-Stream} & \multicolumn{6}{c}{Single-Stream} \\
    %\midrule
    {} & \multicolumn{2}{c|}{\vilbert} & \multicolumn{2}{c|}{\lxmert*} & \multicolumn{2}{c|}{\uniter*} & \multicolumn{2}{c|}{\visualbert*} & \multicolumn{2}{c|}{\vlbert} & \multicolumn{2}{c}{BLIP} \\
    \midrule
    {} & \volta & \hnc & \volta & \hnc & \volta & \hnc & \volta & \hnc & \volta & \hnc & ITM & \hnc \\
    \midrule
    Accuracy   & 55.77 & 55.97 & 53.48 & 55.45 & 55.28 & 56.70 & 53.51 & 56.85 & 55.62 & 55.96 & 57.38 & \textbf{57.73} \\
    \bottomrule
    \end{tabular}
    }
\caption{
Results on the \gls{gqa} \cite{gqa} testdev split. Results are statistically significant*.}
\label{tab:results-gqa-complete}
\end{table*}

\subsection{Caption Generation Settings}
\label{subsec:cpt-gen-setting}
As mentioned in Section \ref{sec:datasets}, we implemented several heuristics to avoid ambiguity and potential noise in our caption generation. We now detail what these heuristics are and how they were implemented.

\paragraph{Ambiguity}
In many caption types, we only address localized cross-modal mismatches by leveraging subgraphs and do not take the global context of a scene into account. This results in ambiguity in entity grounding, especially when multiple instances of the same object class are present in the image. Additionally, scene graphs contain spatial relation annotations between entities and background objects such as \emph{sky} or \emph{field} that typically cover a large area in the scene. This causes ambiguity in captions as the exact spatial relation between them is hard to determine even for humans. Following heuristics are applied to reduce such ambiguities in captions (automatically created as well as human-annotated):
\begin{itemize}
    \item A caption should not refer to multiple instances of the same entity class to avoid ambiguity in terms of entity grounding.
    \item A caption should not refer to a spatial relation between two body parts since such a caption is unnatural as well as error-prone due to multiple instances of body parts in scenes.
    \item A caption should not refer to a spatial relation between an entity and an object typically covering a large area in scenes, i.e.,~typical background objects.
\end{itemize}

\paragraph{Clean vs. noisy}
In our \textbf{clean} setting, we filter out all the values that our noisy spatial relation detection algorithm tags as \emph{noisy}. The way this works is:
\begin{enumerate}
    \item The algorithm gets a triple (subject, relation, object) and a marker as to which value in the tuple should be replaced with a foil.
    \item All the candidate replacement values are collected in a list. This also follows a set of heuristics which we discuss later.
    \item We then compare the bounding boxes of the subject and the object, and decide whether the spatial relation is correct between these visual objects.
    \item If we determine that the given relation is incorrect, we remove this item from the list of candidates.
\end{enumerate}
In the \textbf{noisy} setting, we do not filter out these potentially noisy candidates.

\paragraph{Strict vs. relaxed sampling}
There are several ways of sampling foils for a given tuple. The simplest way would be to sample from all the words in the vocabulary in the same POS tag category, i.e.,~sample from the set of nouns in the vocabulary for a given noun, e.g.,~sample a \underline{shoe} for \underline{cat}. However, as it quickly becomes obvious, this approach has several potential issues. One issue, for example, is that we might end up with nonsensical captions containing an object an unsuitable attribute, e.g.,~\emph{the ground is scrambled} (see \ref{fig:scrambled-ground}.). Also, since the scene graphs contain non-spatial relations, we might accidentally create captions that violate object affordances, e.g.,~\emph{a table is eating a boy}. Thus, it is important to follow an \textbf{informed sampling strategy}. To achieve this, we created look-up tables allowing us to sample a foil that does not result in a nonsensical caption. For (attribute, object), (subject, predicate), and (predicate, object) pairs we aggregate the information in the ground-truth scene graphs and save them as look-up tables. Additionally, we annotated attribute clusters that group similar attributes into buckets for us to sample values from. Using these look-up tables, we provide two negative value sampling strategies for generating hard negative captions: \textbf{(a) relaxed} and \textbf{(b) strict}. 

Our \textbf{relaxed} setting allows sampling from a \emph{probable} set of values such that we allow sampling a negative attribute from the attribute class of the positive one; and for the (subject, predicate, object) triples, we sample from the union of the (subject, predicate) and (predicate, object) pairs. This type of sampling makes the assumption of: given that an object co-occurs with a similar attribute or that a predicate with a subject and an object on different accounts, although an exact tuple might not co-occur in the dataset, this does not mean that such a co-occurrence is unlikely. This increases the variability of the captions but can also result in erroneous cases because neither the attribute clusters are robust (see caption 2 in Figure \ref{fig:cats}.) nor the assumption always holds: if (subject, predicate) and (predicate, object), then (subject, predicate, object), e.g.,~(dog, drinks) and (drinks, beer) does not guarantee (dog, drinks, beer).

In \textbf{strict} setting, we only allow sampling from the look-up tables directly meaning that the exact co-occurrence exists in the ground-truth scene graphs. This results in a highly strict constraint as we essentially limit the likely negative candidates to the ones that co-occur in the dataset. Nonetheless, by doing so, we minimize the number of nonsensical captions. 

In all our experiments, we used the captions generated using the \textbf{clean and strict} setting.

\paragraph{Balancing the comparative quantifiers in captions}
In order to prevent models from attending to linguistic signals for a prediction shortcut, comparative quantifiers are equally used in the positive and the negative caption types.

\paragraph{Balancing the existence and nonexistence in existence-based captions}
Same as above, to avoid shortcuts, \emph{no} and \emph{at least one}, i.e.,~(non)existence of entities, in positive and negative captions are balanced.

\subsection{Qualitative Analysis}
\label{subsec:qualitative-analysis}
\begin{table*}
    \centering
    \resizebox{0.85\textwidth}{!}{
    \begin{tabular}{lllll}
    \toprule
    {} & Clean Strict & Clean Relaxed & Noisy Strict & Noisy Relaxed \\
    \midrule 
    attribute                   &         62.0 &          \textbf{65.2} &         62.3 &          \textbf{65.2} \\ 
    attribute\_relation         &         60.4 &          \textbf{63.3} &         60.3 &          \textbf{63.3} \\
    relation                    &         58.0 &          59.6 &         57.7 &          \textbf{60.1} \\
    relation\_attribute         &         63.2 &          64.8 &         62.9 &          \textbf{65.2} \\
    object\_count               &         65.5 &          65.6 &         \textbf{65.7} &          65.4 \\
    object\_compare\_count      &         61.3 &          \textbf{61.4} &         \textbf{61.4} &          60.8 \\
    verify\_object\_attribute   &         \textbf{55.5} &          55.2 &         55.0 &          55.3 \\ verify\_object\_relation    &         54.1 &          54.2 &         54.0 &          54.0 \\
    AND\_logic\_attribute       &         \textbf{55.4} &          55.2 &         55.1 &          55.2 \\ AND\_logic\_relation        &         55.0 &          56.2 &         54.6 &          56.3 \\
    XOR\_logic\_attribute       &         \textbf{54.3} &          \textbf{54.3} &         53.8 &          53.0 \\  XOR\_logic\_relation       &         60.6 &          62.7 &         58.6 &          \textbf{63.3} \\
    \textbf{all}                &         57.6 &          58.6 &         57.4 &          \textbf{58.7} \\
    \bottomrule
    \end{tabular}
    }
    \caption{Language Model results on \gls{hnc} validation set. The models are trained and evaluated on data obtained from the same setting.}
    \label{tab:lm-val-set}
\end{table*}

\begin{table*}
    \centering
    \resizebox{0.85\textwidth}{!}{
    \begin{tabular}{lllll}
    \toprule
    {} & Clean Strict & Clean Relaxed & Noisy Strict & Noisy Relaxed \\
    \midrule 
    attribute                   &         55.9 &          \textbf{60.4} &         55.4 &          58.4 \\ 
    attribute\_relation         &         51.0 &          \textbf{54.5} &         52.5 &          54.0 \\
    relation                    &         \textbf{56.0} &          55.8 &         54.3 &          54.5 \\
    relation\_attribute         &         53.5 &          54.7 &         \textbf{55.0} &          53.5 \\
    object\_count               &         \textbf{55.0} &          52.5 &         \textbf{55.0} &          54.0 \\
    object\_compare\_count      &         53.5 &          55.5 &         \textbf{56.5} &          54.5 \\
    verify\_object\_attribute   &         \textbf{51.5} &          48.5 &         48.5 &          \textbf{51.5} \\ verify\_object\_relation    &         \textbf{54.0} &          52.5 &         53.0 &          53.5 \\
    AND\_logic\_attribute       &         52.0 &          51.5 &         \textbf{54.0} &          50.5 \\ AND\_logic\_relation        &         \textbf{50.0} &          48.5 &         \textbf{50.0} &          48.5 \\
    XOR\_logic\_attribute       &         48.0 &          48.5 &         \textbf{50.0} &          48.0 \\  XOR\_logic\_relation        &         49.5 &          52.5 &         49.5 &          \textbf{56.0} \\
    \textbf{all}                &         53.1 &          \textbf{53.5} &         53.3 &          53.3 \\
    \bottomrule
    \end{tabular}
    }
    \caption{Language Model results on \gls{hnc} test set. The models are trained on different settings and evaluated on the human-annotated test set.}
    \label{tab:lm-test-set}
\end{table*}

\begin{table*}[t]
    \centering
    \scriptsize
    \resizebox{0.95\textwidth}{!}{
    \begin{tabular}{@{~}l@{~}ll|ll|ll|ll|lll}
    \toprule
    {} & \multicolumn{4}{c|}{Dual-Stream} & \multicolumn{6}{c}{Single-Stream} \\
    \midrule
    {} & \multicolumn{2}{c|}{\vilbert} & \multicolumn{2}{c|}{\lxmert} & \multicolumn{2}{c|}{\uniter} & \multicolumn{2}{c|}{\visualbert} & \multicolumn{2}{c}{\vlbert} \\
    \midrule
    {} & \volta & \hnc & \volta & \hnc & \volta & \hnc & \volta & \hnc & \volta & \hnc \\
    \midrule
    attribute          & 50.5 & 57.9 & 52.0 & 56.9 & 54.0 & 58.4 & 52.0 & 54.5 & 50.0 & \textbf{61.9} \\
    attribute\_rel     & 49.5 & \textbf{56.0} & 52.5 & 54.5 & 52.0 & 53.5 & 49.5 & 54.0 & 50.0 & 52.5 \\
    relation           & 49.0 & \textbf{54.8} & 49.3 & 54.5 & 49.2 & 53.0 & 50.0 & 54.0 & 50.0 & 53.3 \\
    relation\_attr     & 50.9 & 55.5 & 50.9 & 56.2 & 49.6 & 57.1 & 49.9 & 54.7 & 50.3 & \textbf{57.8} \\
    obj\_count         & 49.5 & \textbf{65.0} & 45.0 & 46.5 & 51.0 & 61.0 & 51.5 & 64.0 & 50.5 & 58.0 \\
    obj\_comp\_count   & 50.5 & 51.5 & 49.5 & 53.0 & 48.0 & 53.0 & 50.5 & 52.5 & 49.0 & \textbf{53.5} \\
    verify\_obj\_attr  & \textbf{52.5} & 42.0 & 51.5 & 44.5 & 46.0 & 46.0 & 47.5 & 46.5 & 50.0 & 50.0 \\
    verify\_obj\_rel   & 50.5 & \textbf{54.5} & 51.0 & 53.5 & 50.0 & 51.0 & 50.0 & 53.0 & 50.0 & 52.0 \\
    AND\_logic\_attr   & 51.5 & \textbf{59.0} & 49.0 & 50.0 & 49.5 & 51.0 & 51.0 & 52.0 & 51.0 & 57.0 \\
    AND\_logic\_rel    & 50.0 & \textbf{54.0} & 48.5 & 53.5 & 49.0 & 52.5 & 50.0 & 49.0 & 50.0 & 49.5 \\
    XOR\_logic\_attr   & 49.5 & \textbf{53.0} & 50.5 & 49.0 & 50.5 & 51.0 & 49.0 & 49.5 & 50.0 & 50.5 \\
    XOR\_logic\_rel    & 50.0 & 57.5 & 49.5 & \textbf{60.0} & 49.5 & 54.0 & 50.0 & 57.0 & 50.0 & 59.0 \\
    \textbf{all}       & 50.2 & \textbf{55.1} & 50.0 & 53.3 & 50.0 & 53.8 & 50.1 & 53.6 & 48.1 & 50.1 \\ 
    \bottomrule
    \end{tabular}
    }
\caption{\protect Binary classification accuracy on \gls{hnc} test set under blind evaluation.} 
\label{tab:hnc-test-blind}
\end{table*}

\subsubsection{Dataset Generation Process}
\label{subsec:dataset-gen}

\paragraph{Refinements in sampling methods}\label{pa:dataset-gen-sampling}
In our first iteration of the sampling implementation, we started with a single constraint, i.e.,~the negative value (object, attribute, relation) must be sampled from within the scene. This, however, results in a strong linguistic bias as there is no mechanism that ensures the sensibility of the generated caption. This resulted in captions like \emph{the table is sleeping}, or \emph{the man is eating a couch} which then gave us \gls{lm} accuracies of approx. $70\%$ on the validation set. This is highly undesirable as the entailment between an image and its caption can be predicted simply by assessing the caption's sensibility. 

In our next iteration of sampling from look-up tables in the \textbf{relaxed} setting, we were able to reduce the \gls{lm} accuracies down to approx. $66\%$. This setting helps us avoid creating captions such as \emph{the man is eating a couch} as the object \underline{eating} does not occur together with \underline{couch} in the ground-truth scene graphs. Note that, at this time, we are using the look-up tables, but we are still sampling uniformly. This uniform sampling turned out to be highly problematic as the word distributions between the positive and the negative captions were too dissimilar resulting in shortcut predictions.  The reason is that co-occurrences of visual concepts in the ground-truth \gls{gqa} scene graphs are highly imbalanced. For example, \emph{to the left of} and \emph{to the right of} are the most common predicates in the dataset. When we uniformly sample from the above-mentioned look-up tables, we create a distributional bias between the positive and the negative caption sets (see subplots \emph{(a) \& (b)} of Figure \ref{fig:rel_distributions} for the relation distribution of the captions from an early iteration.). Thus, we extracted word co-occurrence statistics from the ground-truth scene graphs and sampled from the look-up tables following these distributions (see subplots \emph{(c) \& (d)} of Figure \ref{fig:rel_distributions} for the relation distribution in our final captions.), which helped us reduce the \gls{lm} accuracies down to approx. $58\%$. To reduce the linguistic bias even further, we implemented \textbf{strict} sampling which we detailed in Section \ref{subsec:cpt-gen-setting}. With this sampling strategy, we are able to reduce the \gls{lm} accuracies down to approx. $57\%$ (see Tab.\ref{tab:lm-val-set}).

Table \ref{tab:lm-val-set} shows the \gls{lm} accuracies\footnote{The higher the accuracy, the more biased is the dataset.} on the final versions of the \gls{hnc} validation sets. According to these numbers, some of the caption types contain more bias than the others, e.g.,~\textbf{attribute}, \textbf{attribute\_relation}, \textbf{relation}, \textbf{relation\_attribute}, \textbf{object\_count}, \textbf{object\_compare\_count}, \textbf{XOR\_logic\_relation} all have accuracies $\gtrsim60\%$. For example, the model achieves approx. $65\%$ accuracy on the validation split in \textbf{object\_count} type (approx. $61\%$ in \textbf{object\_compare\_count}). We attribute this to a combination of dataset and world-priors biases which is common in datasets of real-world images.

Note that \gls{lm} accuracies are a simple proxy we use to measure the linguistic bias in the textual modality without the presence of the visual modality. Thus, we believe that none of the methods is ideal, and the choice of the sampling strategy might depend on the use case. 

\begin{figure*}
     \centering
     \begin{subfigure}[b]{0.49\textwidth}
         \centering
         \includegraphics[width=\textwidth]{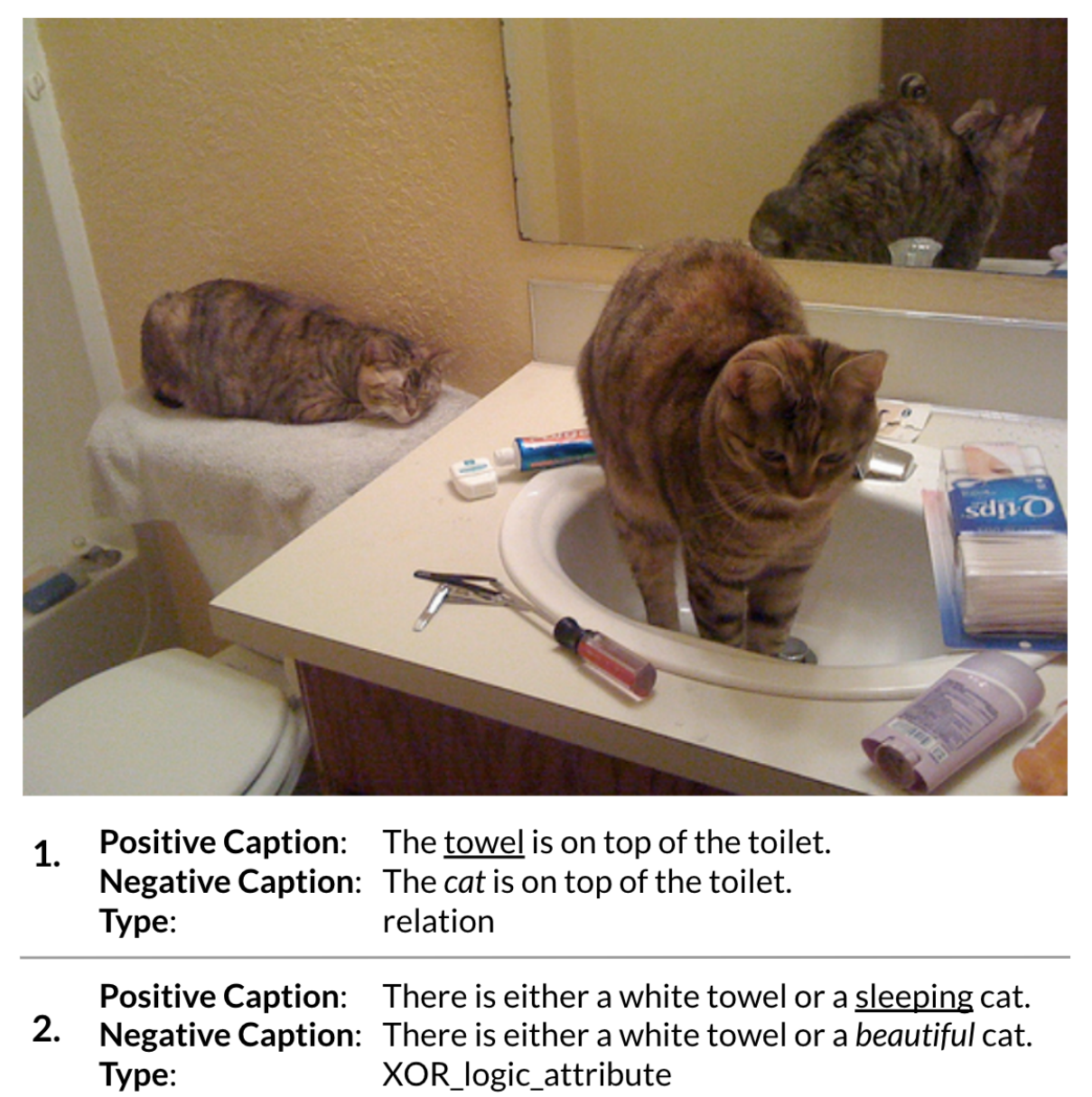}
         \caption{}
         \label{fig:cats}
     \end{subfigure}
     \begin{subfigure}[b]{0.49\textwidth}
         \centering
         \includegraphics[width=\textwidth]{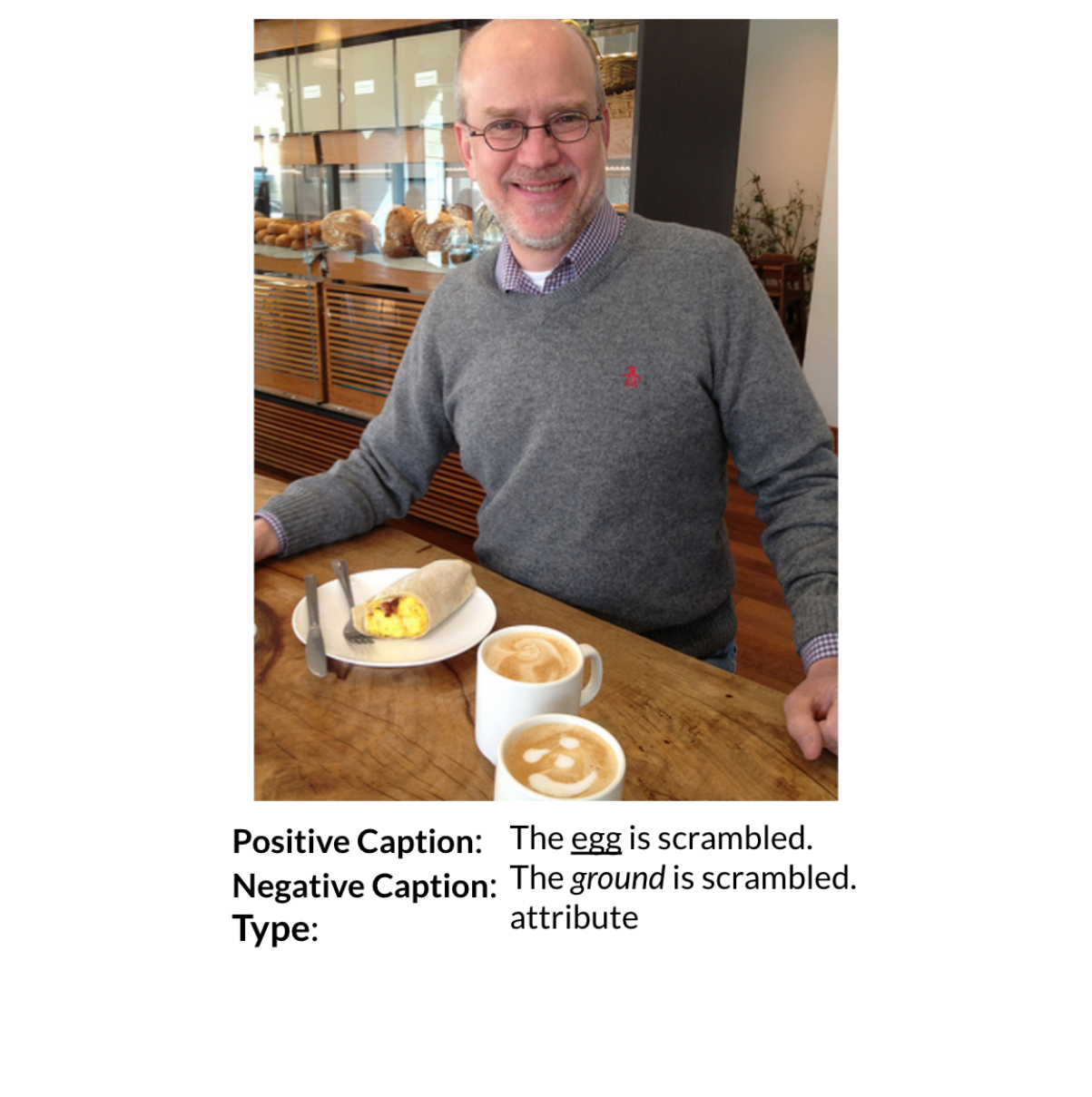}
         \caption{}
        \label{fig:scrambled-ground}
     \end{subfigure}
     \caption{\textbf{(a)} The resulting negative captions do not contradict the image; thus, they are false negatives. \textbf{Negative caption 1} contains a noisy spatial relation, \textbf{negative caption 2} contains an attribute similar to the attribute in the positive caption but not contradictory to the image. \textbf{(b)} The sampled noun ground with the attribute \textit{``scrambled''} creates a nonsensical caption.}
     \label{fig:cpt-gen-examples}
\end{figure*}

\paragraph{Noisy spatial relations}\label{pa:dataset-gen-spatial}  Our qualitative iterative analysis revealed that, due to the incomplete nature of the relations in \gls{gqa} scene graphs, our \emph{noisy} setting results in many noisy hard negative captions in that the values we sample as foils do not contradict the image (see caption 1 in Figure \ref{fig:cats}) However, this is not detectable simply by looking at the \gls{lm} accuracies as the captions are not nonsensical. Thus, between the \emph{clean} and the \emph{noisy} settings, there does not seem to be a great deal of difference for the \gls{lm} which is expected as the sensibility of the captions are not directly affected by the correctness of objects' spatial relations in the visual scene, e.g.,~a bus driver can be \underline{inside} or the \underline{next to} a bus.

\subsubsection{Analysis of the Human-Annotated Test Set}
\label{subsec:modality-ablation}

We evaluated the \gls{lm} trained on \gls{hnc} captions to quantify the pure linguistic bias that might be present in our human-annotated test set. Ideally, \gls{lm} should perform at the random baseline level, i.e.,~$50\%$ accuracy. In our \textbf{clean and strict} setting, the model achieves an average accuracy of $53.1\%$ which suggests the presence of \emph{some} bias. This might be due to the domain size in \gls{gqa} images. Thus, no matter if created automatically or annotated by humans, such statistical biases caused by the domain size are hard to mitigate.

Table \ref{tab:hnc-test-blind} contrasts the accuracies of models trained on \gls{hnc} image--text pairs\footnote{The models are trained on the \emph{clean-strict} version.} with the \gls{volta} models evaluated on the text-only modality of the human-annotated test set (see \ref{subsec:blind-details} for the implementation details). Previously, we discussed biases in our dataset. With these results, our aim is to draw attention to the biases in the pretrained \gls{vl} models. As also briefly mentioned in Section \ref{subsec:main-modality-ablation}, we might violate world-priors in \gls{vl} models by creating negative captions that are possible but might not be probable according to their worldview, e.g.,~the leaves might be more likely to be \underline{green} or \underline{yellow} than \underline{red} or \underline{brown}, although red or brown leaves are not impossible. Moreover, due to the size of the \gls{gqa} images, it is unlikely that the dataset is an accurate sample of the world, i.e.,~although we might have images showing \emph{a man eating pizza} and \emph{a woman eating pasta}, this does not mean that the men do not eat pasta or the other way around.

\subsubsection{Downstream Tasks}
\begin{figure*}
    \centering
    \resizebox{1.0\textwidth}{!}{
        \includegraphics{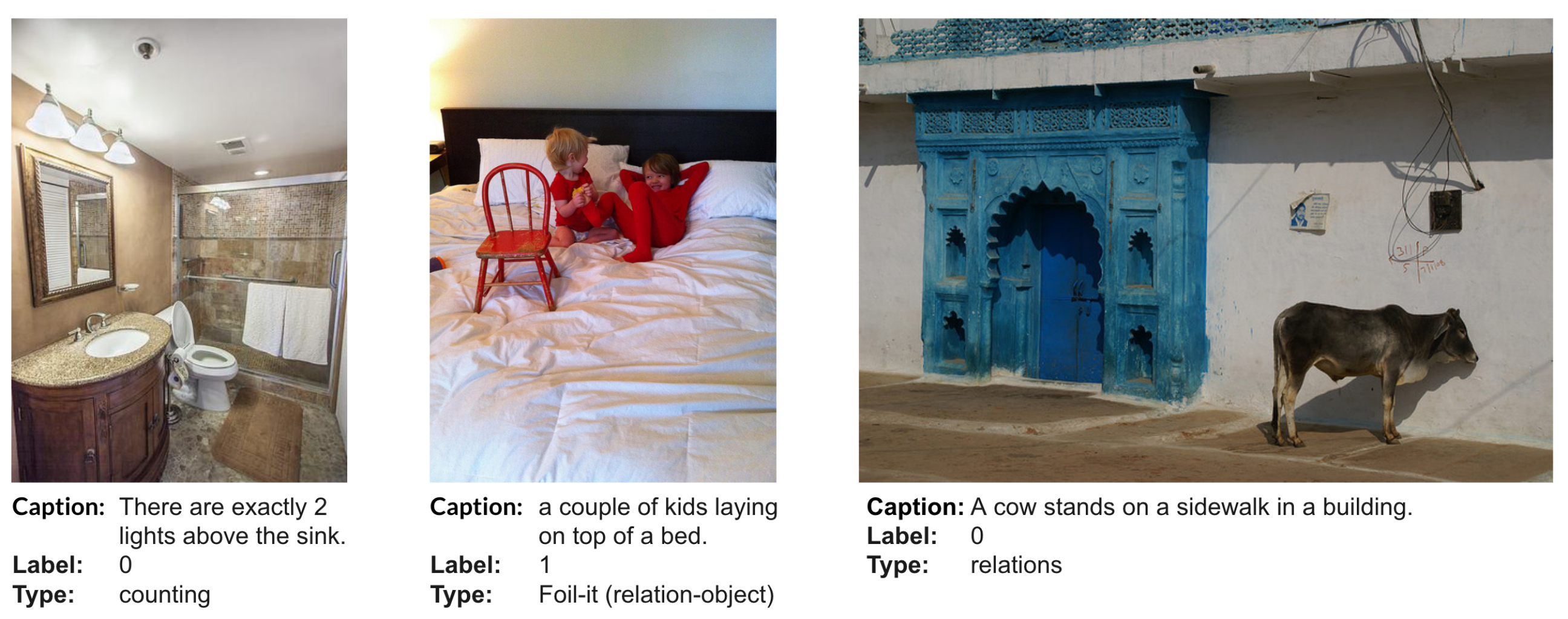}
    }
    \caption{Example cases where all the \gls{volta} models failed while our models predicted the correct entailment.}
    \label{fig:valse-examples}
\end{figure*}

\paragraph{\gls{valse}}
\label{pa:valse_analysis}
In Figure \ref{fig:valse-examples}, we display some examples where all our models predicted the correct entailment between the image and the caption that were predicted incorrectly by all the models initialized from the \gls{volta} checkpoints. 
As also indicated by the quantitative results, we observed significant improvement in all the models regarding certain types of foils, which we discuss briefly in the following. 

Our models predict correct entailment in many counting-based captions that were predicted incorrectly by the \gls{volta} models. Our qualitative analysis revealed that this is especially the case when the foiled count is small and close to the original count. 
Furthermore, in many of our hard negative captions, we swap grammatical subjects (agent, actant) or objects (patient, theme, experiencer) of the captions with a foil. This seems to help models ground the correct visual object in the image and not just predict entailment by assessing the plausibility of the caption. 
We also observe improvements in spatial relation grounding which is expected as our dataset contains many captions that specifically foil this information. In some examples, where \gls{valse} foils the action in the caption, our models perform better as well. 
This might mean that the correct grounding of the subjects and the objects in captions might have a positive effect on the grounding of the action in the visual scene. 
However, since the \gls{gqa} scene graphs do not readily provide many actions, we do not see a big improvement in this type. 

\begin{figure*}
    \centering
    \resizebox{1.0\textwidth}{!}{
        \includegraphics{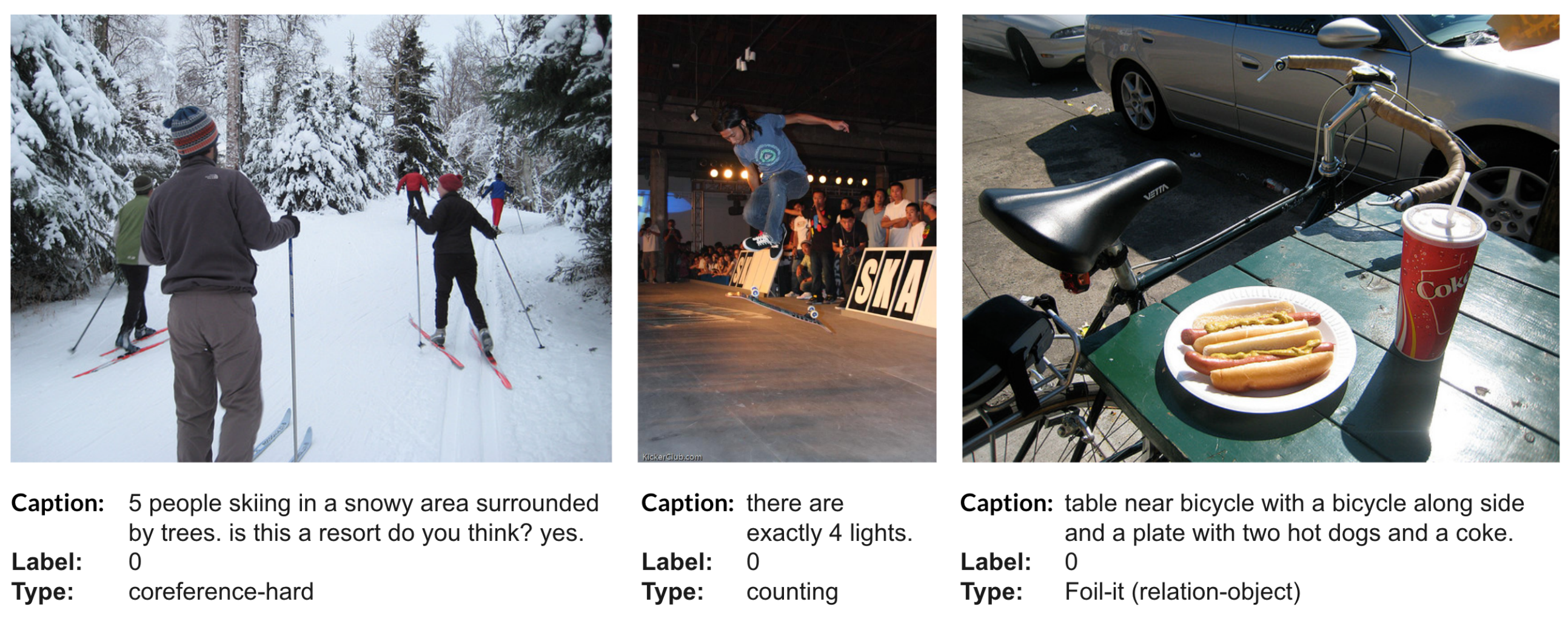}
    }
    \caption{Example cases where all our models failed while the \gls{volta} models predicted the correct entailment.}
    \label{fig:valse-examples-incorrect}
\end{figure*}

We also observed some failure cases where the previously correct predictions were predicted incorrectly by all our models (see Figure \ref{fig:valse-examples-incorrect}). This mainly occurred in foil types that we do not cover in our hard negative caption generation, e.g.,~coreference (see the left example in Figure \ref{fig:valse-examples-incorrect}), plurals and non-spatial relations. However, lack of coverage is not the only place where we observe such behavior. For example, some counting-hard captions that were predicted correctly by \gls{volta} models ended up being predicted incorrectly by all our models (see the middle example in Figure \ref{fig:valse-examples-incorrect}). This might be due to the imbalanced object counts in the captions. We chose to follow the ground-truth scene graph distributions which inherently contain some bias on a compositional level as discussed in Section \ref{subsec:dataset-gen}. The implication of this is that our positive (also hard negative) captions might never have certain combinations of concepts compositionally co-occur in the same caption, i.e.,~while we might have captions that contain one, two, three, or four elephants; we might never have a caption with five elephants in the positive captions if such a scene graph does not exist in the \gls{gqa} dataset. 

Additionally, we found that some of the foiled instances incorrectly predicted by \gls{hnc} models are ambiguous; e.g.,~in the right example of Figure \ref{fig:valse-examples-incorrect}, the foil (bicycle) for the correct object (car) is also near the table.

\begin{figure*}
    \centering
    \resizebox{\textwidth}{!}{
    \includegraphics{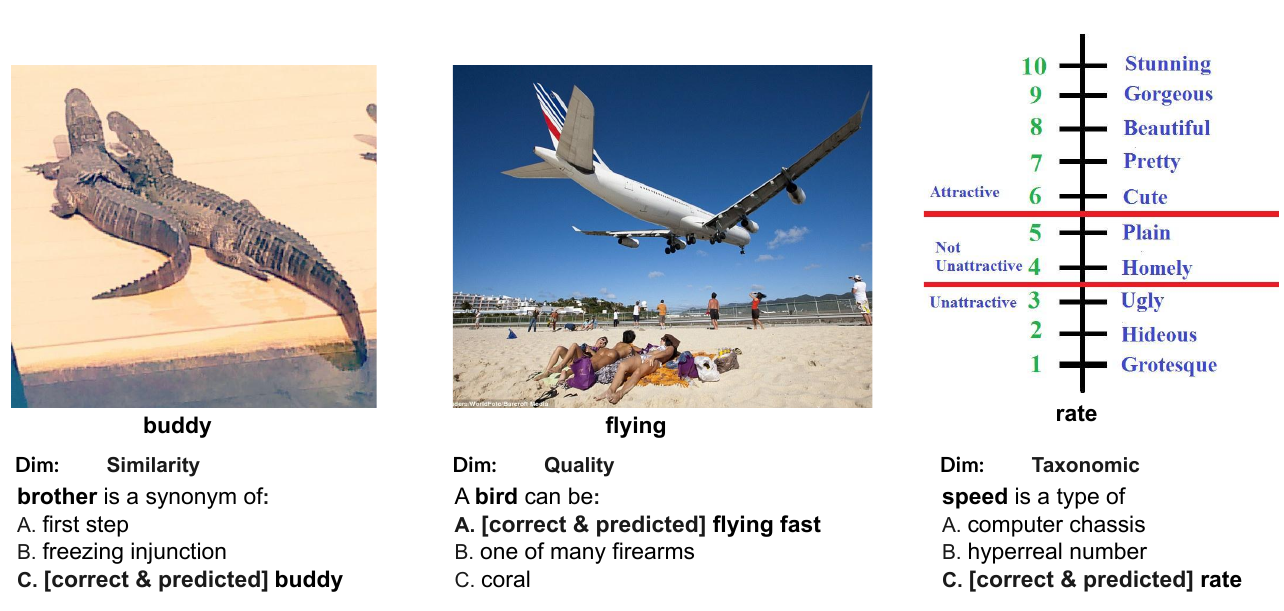}
    }
    \caption{Example cases where our \gls{hnc} single-stream models succeed under noisy visual input scenarios, i.e.,~a modality mismatch between the textual token in the prompt and the image retrieved based on the correct textual choice, e.g., the word \textbf{bird} and the image \textbf{flying}.}
    \label{fig:cpt-examples-hnc-correct}
\end{figure*}

\begin{figure*}
    \centering
    \resizebox{0.5\textwidth}{!}{
    \includegraphics{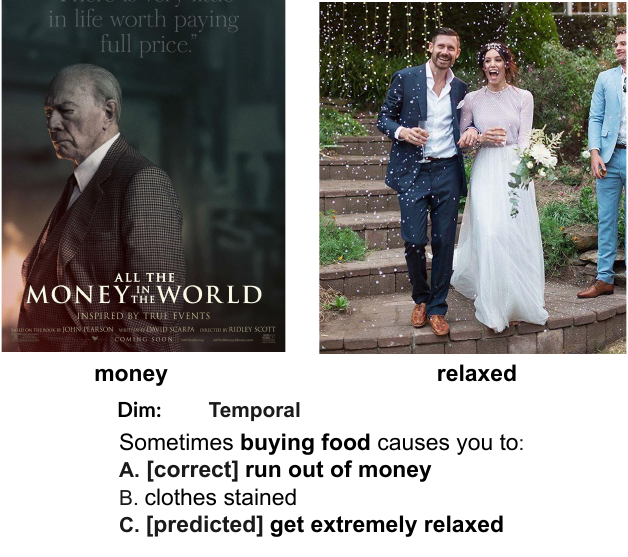}
    }
    \caption{A failure case of \gls{hnc} dual-stream models on the temporal dimension.}
    \label{fig:cpt-examples-hnc-dual-incorrect-temporal}
\end{figure*}

\begin{figure*}
    \centering
    \resizebox{0.6\textwidth}{!}{
    \includegraphics{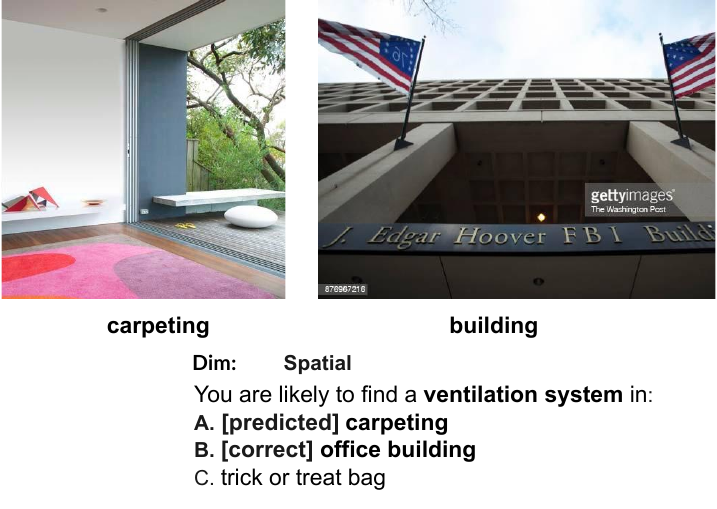}
    }
    \caption{A failure case of \gls{hnc} dual-stream models on the spatial dimension.}
    \label{fig:cpt-examples-hnc-dual-incorrect-spatial}
\end{figure*}

\paragraph{\gls{cpt}}
\label{pa:cpt_analysis}
Each instance of \cwwv consists of three natural language statements and a corresponding set of retrieved images,~$T_i=(Q||A_i||V_i)$, \mbox{$i=1,\dots,3$},  where $Q$~is the prompt,~$A_i$ a candidate answer,~$V_i$ is a set of retrieved images for the answer tokens. A model has to determine in a zero-shot manner which of the three statements is true. Specifically, it requires a model to perform \gls{mlm} on the same masked token of the prompt $Q$ in each $T$. The statement that receives the lowest \gls{mlm} loss is considered the model's prediction.

In Figure \ref{fig:cpt-examples-hnc-correct}, we showcase several examples where \gls{hnc} single-stream models successfully handle noisy visual inputs during the inference stage (\gls{volta} single-stream models fail), especially on \textbf{similarity}, \textbf{quality}, and \textbf{taxonomic} dimension. 
We investigate how the visual noisiness in the aforementioned dimensions varies from each other by looking into respective examples.
For \textbf{similarity}, although the extracted image metaphorically captures the answer token, \textit{buddy}, to display a sense of togetherness, there is no human being, but only two crocodiles, in the picture, which creates an entity-level misalignment w.r.t the question token, \textit{brother}, in the prompt. A similar issue is observed for the \textbf{quality} dimension, in which the extracted image for \textit{flying} is conceptually correct, but no \textit{bird}, but only a plane, can be identified in the image. As for \textbf{taxonomic} dimension, we found that general concept words like \textit{rate} could potentially create a modality misalignment issue w.r.t. the question token in the prompt, e.g.,~\textit{speed} because \textit{rate} could also be a unit to measure attractiveness in this case.
These cases exemplify the difficulty of \gls{cpt} task that might lead \gls{vl} models to pick a wrong prediction in the presence of conceptually correct, but not-strictly-aligned, visual inputs. 
However, since \gls{hnc} single-stream models are pre-trained to be aware of fine-grained misalignment, they bypass the limited information provided by the visual modality and robustly resort to the textual modality for performing inference. 
The effectiveness does not generalize to other dimensions such as \textbf{temporal} and \textbf{spatial} as exemplified in Figure \ref{fig:cpt-examples-hnc-dual-incorrect-temporal} and Figure \ref{fig:cpt-examples-hnc-dual-incorrect-spatial} respectively. It is notable that \gls{hnc} dual-stream models suffer stronger from a performance decrease than the single-stream counterparts.
By inspecting the failure case of \textbf{temporal} made by \gls{hnc} dual-stream, it is clear that the wrong prediction could easily occur due to the natural misalignment of the temporal orders between the question token, \textit{buying food}, in the prompt and the answer token, \textit{run out of money}. Therefore, the resulting retrieved image is naturally not corresponding. In the example here, we observe \gls{hnc} dual-streams select the choice, \textit{get extremely relaxed}. The reason behind this could be that there are glasses, hyponyms of \textit{food}, existing in the \textit{relaxed} picture. 
With respect to the failure case of \textit{spatial} dimension, again, we see that \gls{hnc} dual streams are subject to slight modality non-correspondence. The image extracted for the correct answer token, \textit{building} capture the external view of a \textit{building}; whereas the image for the wrongly picked answer token, \textit{carpeting}, is photographed inside a house.  

\subsection{Statistical Test}
\label{subsec:statistical-test}
To determine whether one model significantly outperforms the other one, we resort to paired student’s t-test \cite{Fisher1949-qa} with the threshold of p < 0.05 to be significantly outperforming. Since the t-test assumes a normal distribution, we also test the normality of model prediction with the method of Anderson-Darling \cite{Anderson1954-vq}.

\subsection{Dataset Statistics}
\begin{figure}[h]
    \resizebox{0.99\linewidth}{!}{
    \centering
    \includegraphics{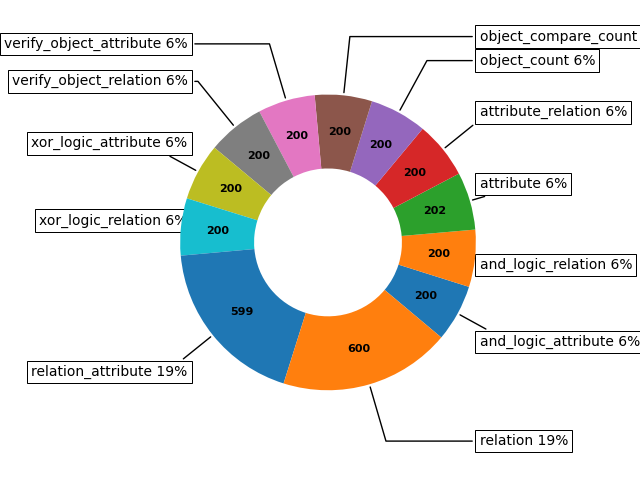}
    }
    \caption{Test set caption type distribution.}
    \label{fig:test_set_cpt_distribution}
\end{figure}

Figure \ref{fig:test_set_cpt_distribution} contains the distributions for the human annotated test set.
The total number of each caption type as well as the relative percentage values are displayed.
The test set contains exactly $100$ annotated images.

\begin{figure*}
     \centering
     \begin{subfigure}[b]{0.42\textwidth}
         \centering
         \includegraphics[width=\textwidth]{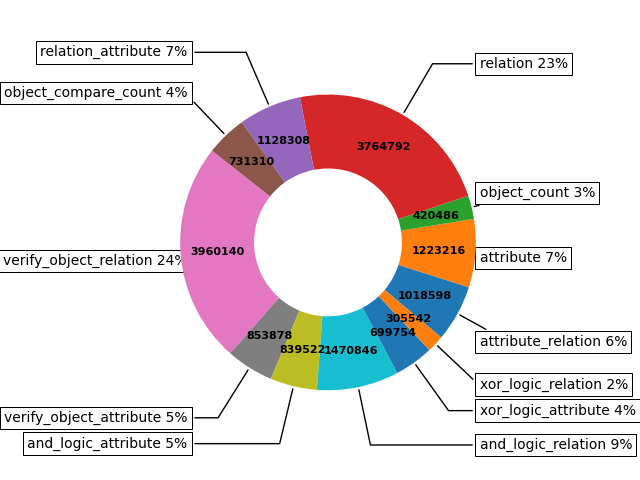}
         \caption{Clean strict.}
     \end{subfigure}
     \begin{subfigure}[b]{0.42\textwidth}
         \centering
         \includegraphics[width=\textwidth]{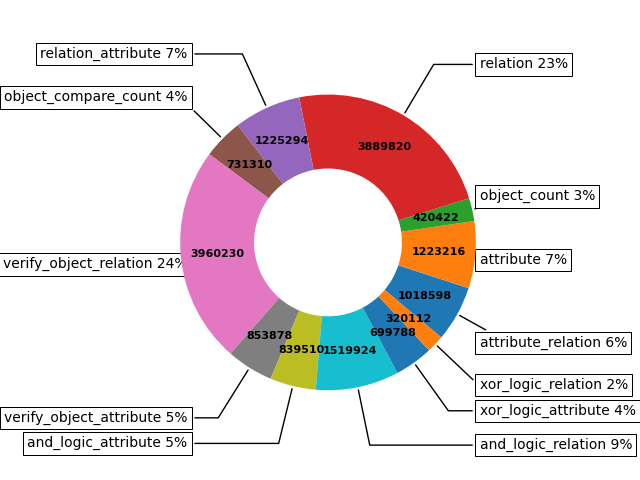}
         \caption{Noisy strict.}
     \end{subfigure}
     \begin{subfigure}[b]{0.42\textwidth}
         \centering
         \includegraphics[width=\textwidth]{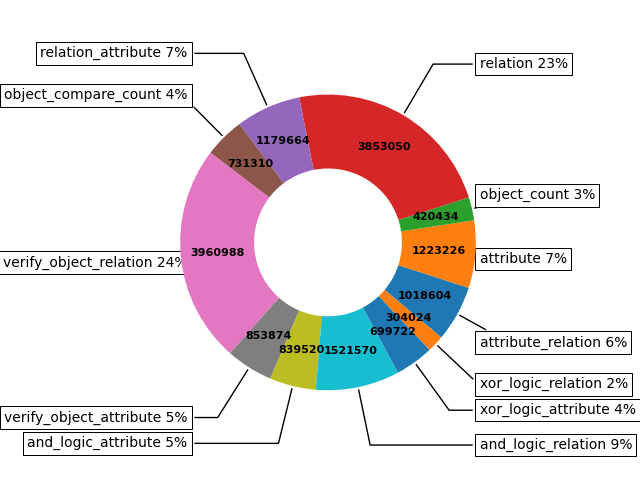}
         \caption{Clean relaxed.}
     \end{subfigure}
     \begin{subfigure}[b]{0.42\textwidth}
         \centering
         \includegraphics[width=\textwidth]{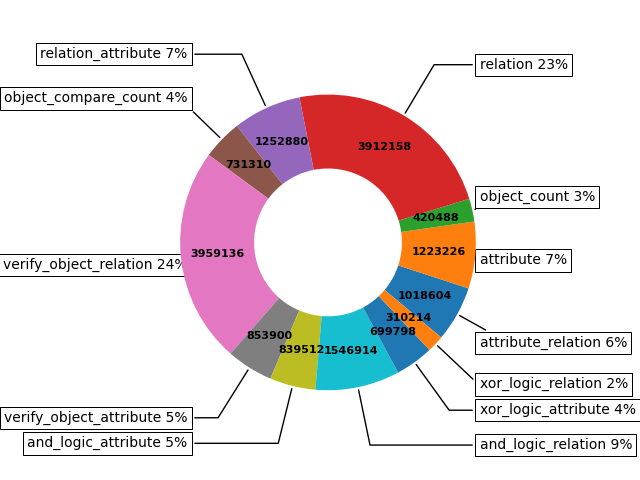}
         \caption{Noisy relaxed.}
     \end{subfigure}
    \caption{Training split variation distributions.}
    \label{fig:dec7_train_cpt_distributions}
\end{figure*}

\begin{figure*}
     \centering
     \begin{subfigure}[b]{0.42\textwidth}
         \centering
         \includegraphics[width=\textwidth]{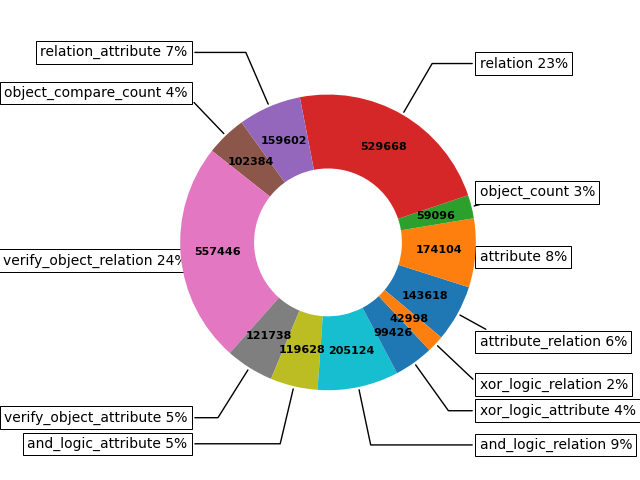}
         \caption{Clean strict.}
     \end{subfigure}
     \begin{subfigure}[b]{0.42\textwidth}
         \centering
         \includegraphics[width=\textwidth]{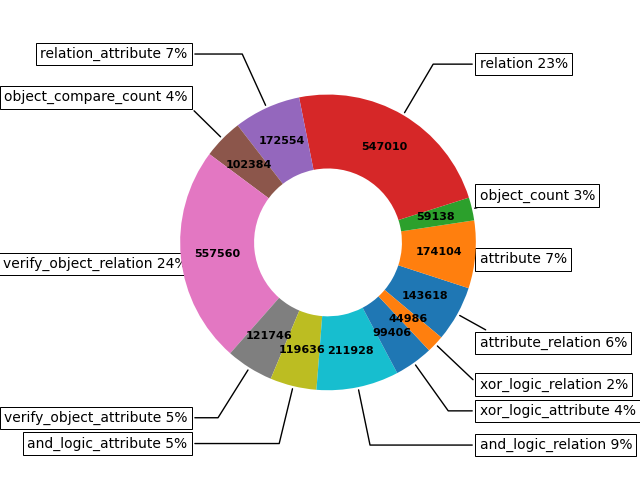}
         \caption{Noisy strict}
     \end{subfigure}
     \begin{subfigure}[b]{0.42\textwidth}
         \centering
         \includegraphics[width=\textwidth]{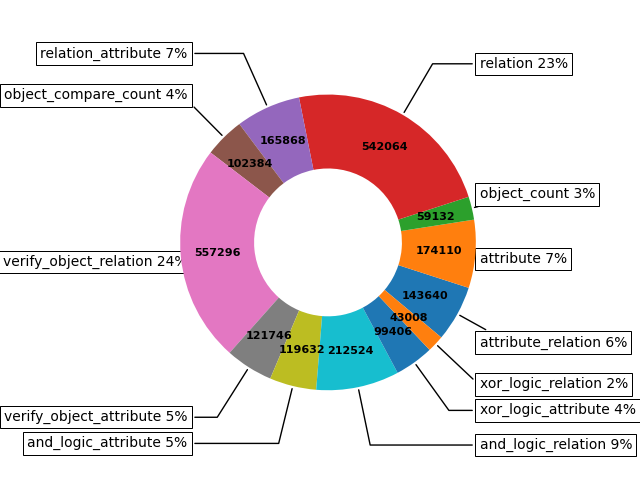}
         \caption{Clean relaxed.}
     \end{subfigure}
     \begin{subfigure}[b]{0.42\textwidth}
         \centering
         \includegraphics[width=\textwidth]{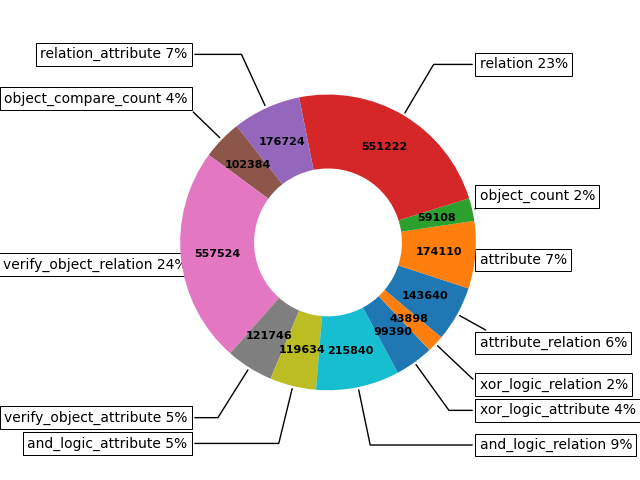}
         \caption{Noisy relaxed.}
     \end{subfigure}
    \caption{Validation split variation distributions.}
    \label{fig:dec7_valid_cpt_distributions}
\end{figure*}

Figure \ref{fig:dec7_train_cpt_distributions} contains the caption type distributions for the training set data w.r.t. the different dataset variations, and Figure \ref{fig:dec7_valid_cpt_distributions} contains the caption type distributions for the validation set.

\begin{figure*}
     \centering
     \begin{subfigure}[b]{0.42\textwidth}
         \centering
         \includegraphics[width=\textwidth]{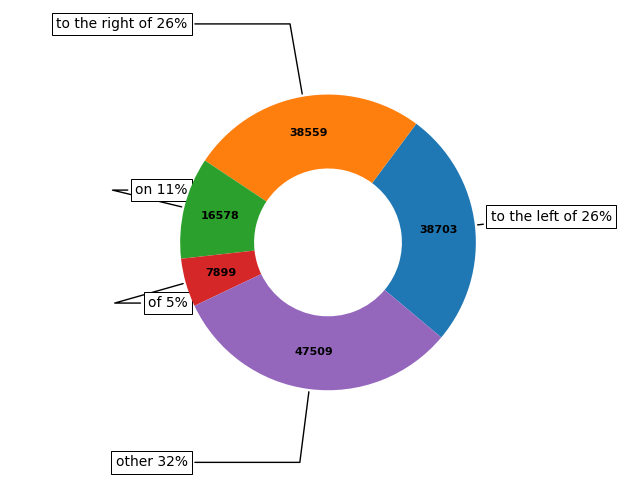}
         \caption{Negative captions validation split. Early iteration.}
         \label{fig:rel_distributions_a}
     \end{subfigure}
     \begin{subfigure}[b]{0.42\textwidth}
         \centering
         \includegraphics[width=\textwidth]{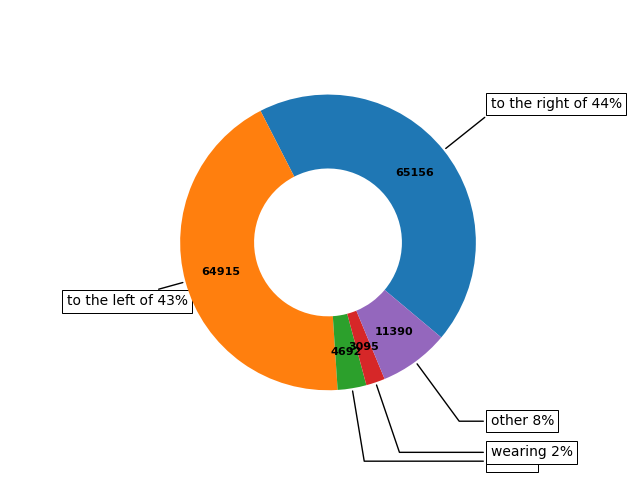}
         \caption{Positive captions validation split. Early iteration.}
         \label{fig:rel_distributions_b}
     \end{subfigure}
     \begin{subfigure}[b]{0.42\textwidth}
         \centering
         \includegraphics[width=\textwidth]{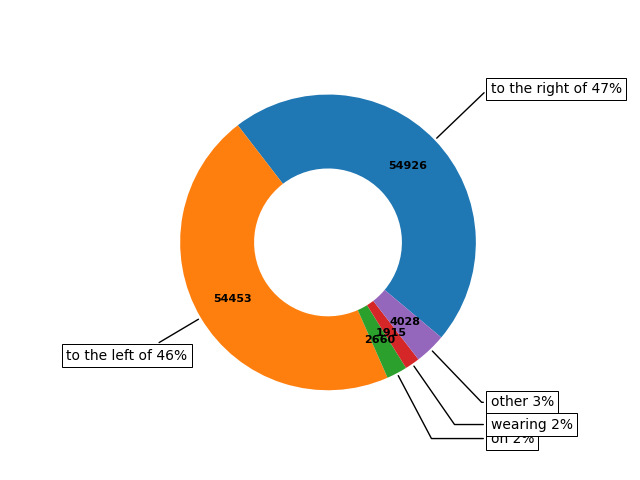}
         \caption{Negative captions validation split. Final iteration.}
         \label{fig:rel_distributions_c}
     \end{subfigure}
     \begin{subfigure}[b]{0.42\textwidth}
         \centering
         \includegraphics[width=\textwidth]{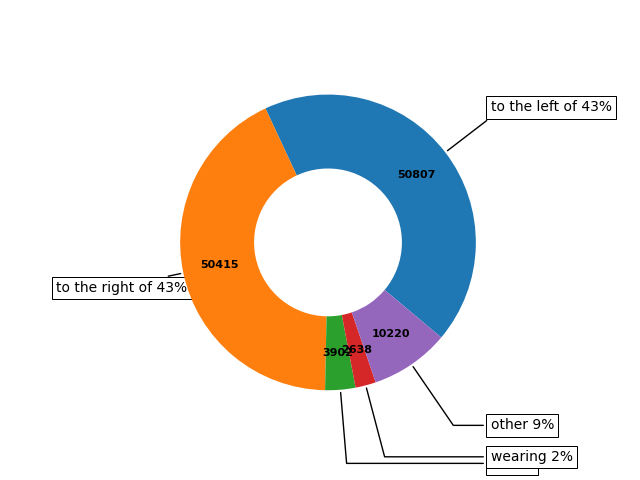}
         \caption{Positive captions validation split. Final iteration.}
         \label{fig:rel_distributions_d}
     \end{subfigure}
    \caption{Relations distributions.}
    \label{fig:rel_distributions}
\end{figure*}

Figure \ref{fig:rel_distributions} displays the relation distributions for the positive and negative captions.
Fig. \ref{fig:rel_distributions_a} and \ref{fig:rel_distributions_b} contain the distributions from earlier iterations.
It is striking to see that the relation distributions in the positive and negative captions are very dissimilar.
Our final state of the caption generation procedure produces similar relation distributions, as can be found in Fig. \ref{fig:rel_distributions_c} and \ref{fig:rel_distributions_d}.
Most prominent are the relations \emph{to the left of} and \emph{to the right of}.
Following different data distributions enables models to easily distinguish between negative and positive captions, which is why we mitigated the gap between iterations.

Table \ref{tab:Splits_Full} contains the exact numbers for each dataset split and variation.
\begin{table*}[hb]
    \scriptsize
    \resizebox{0.99\textwidth}{!}{
    \begin{tabular}{llrrr}
    \toprule
    Split & Variation & Total Amount Cpts & Avg Cpt Len & Avg Cpt Amounts across Types \\
    \midrule
    \multirow{4}{*}{Valid} & Clean Strict  & 2,314,832 & 10.28 & 238.81 \\
                           & Clean Relaxed & 2,340,810 & 10.26 & 241.49 \\
                           & Noisy Strict  & 2,354,070 & 10.27 & 242.86 \\
                           & Noisy Relaxed & 2,365,220 & 10.25 & 244.01 \\
    \midrule
    \multirow{4}{*}{Train} & Clean Strict  & 16,416,392 & 10.29 & 242.10\\
                           & Clean Relaxed & 16,605,986 & 10.27 & 244.90\\
                           & Noisy Strict  & 16,702,102 & 10.29 & 246.32\\
                           & Noisy Relaxed & 16,768,140 & 10.27 & 247.29\\
    \bottomrule
    \end{tabular}
    }
\caption{Statistics of our automatically generated data splits and variations.}
\label{tab:Splits_Full}
\end{table*}

\end{document}